\documentclass[letterpaper, 10 pt, conference]{ieeeconf}

\IEEEoverridecommandlockouts

\usepackage{mathptmx}
\usepackage{helvet}
\usepackage{courier}
\usepackage{type1cm}
\usepackage{makeidx}
\usepackage{graphicx}
\usepackage{multicol}
\usepackage[bottom]{footmisc}

\usepackage{graphics}
\usepackage{amsmath}
\usepackage{amssymb}
\usepackage{graphicx}
\usepackage[caption=false,subrefformat=parens,labelformat=parens]{subfig}
\usepackage{comment}
\usepackage[normalem]{ulem}

\usepackage{color}
\usepackage{url}
\usepackage{multirow}

\title{\LARGE \bf
Efficient Generation of Motion Plans from Attribute-Based Natural Language Instructions Using Dynamic Constraint Mapping
}

\author{Jae Sung Park$^{1}$, Biao Jia$^{2}$, Mohit Bansal$^{1}$, and Dinesh Manocha$^{3}$
\thanks{$^{1}$Department of Computer Science, University of North Carolina at Chapel Hill, USA {\tt\small \{jaesungp,mbansal\}@cs.unc.edu}}%
\thanks{$^{2}$Department of Electrical and Computer Engineering, University of Maryland, USA {\tt\small biao@cs.umd.edu}}%
\thanks{$^{3}$Department of Computer Science and Electrical and Computer Engineering, University of Maryland, USA {\tt\small dm@cs.umd.edu}}%
}

\begin{document}

\maketitle
\thispagestyle{empty}
\pagestyle{empty}

\begin{abstract}
We present an algorithm for combining natural language processing (NLP) and fast robot motion planning to automatically generate robot movements. Our formulation uses a novel concept called Dynamic Constraint Mapping to transform complex, attribute-based natural language instructions into appropriate cost functions and parametric constraints for optimization-based motion planning. We generate a factor graph from natural language instructions called the Dynamic Grounding Graph (DGG), which takes latent parameters into account. The coefficients of this factor graph are learned based on conditional random fields (CRFs) and are used to dynamically generate the constraints for motion planning. We map the cost function directly to the motion parameters of the planner and compute smooth trajectories in dynamic scenes. We highlight the performance of our approach in a simulated environment and via a human interacting with a 7-DOF Fetch robot using intricate language commands including negation, orientation specification, and distance constraints.
\end{abstract}

\section{Introduction}

In the field of human-robot interaction (HRI), natural language has been used as an interface to communicate a human's intent to a robot~\cite{kollar2013generalized,howard2014natural,branavan2009reinforcement,matuszek2010following}. Much of the work in this area is related to specifying simple tasks or commands for robot manipulation, such as picking up and placing objects. As robots are increasingly used in complex scenarios and applications, it is important to develop a new generation of motion planning and robot movement techniques that can respond appropriately to diverse, attribute-based NLP instructions for HRI, e.g., instructions containing negation based phrases or references to position, velocity, and distance constraints. Furthermore, we need efficient techniques to automatically map the NLP instructions to such motion planners.

Humans frequently issue commands that include sentences with orientation-based or negation constraints such as ``put a bottle on the table and keep it upright'' or ``move the knife but don't point it towards people,'' or sentences with velocity-based constraints such as ``move slowly when you get close to a human.'' To generate robot actions and movements in response to such complex natural language instructions, we need to address two kinds of challenges:

\noindent 1. The accurate interpretation of attribute-based natural language instructions and their grounded linguistic semantics, especially considering the environment and the context. For example, a human may say ``move a little to the left'' or ``do not move like this,'' and the robot planner needs to learn the correct interpretation of these commands that include spatial and motion-based adjectives, adverbs, and negation.

\noindent 2. The motion planner needs to generate appropriate trajectories based on these complex natural language instructions. This includes appropriately setting up the motion planning problem  based on different motion constraints (e.g., orientation, velocity, smoothness, and avoidance) and computing smooth and collision-free paths.

\begin{figure}[t]
  \centering
  \subfloat[][]
  {
    \includegraphics[width=0.32\linewidth]{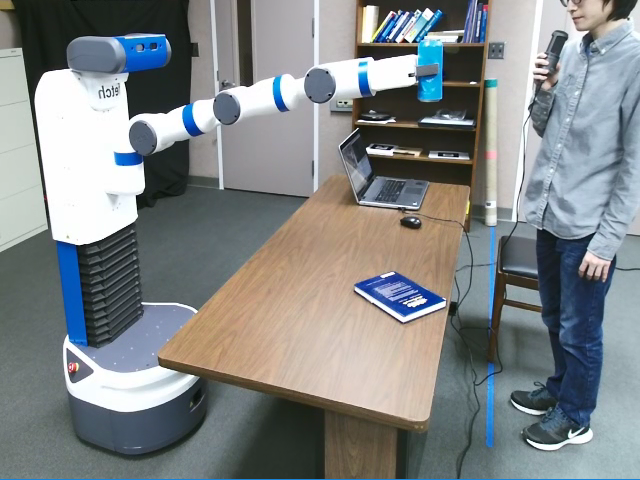}
  }
  \subfloat[][]
  {
    \includegraphics[width=0.32\linewidth]{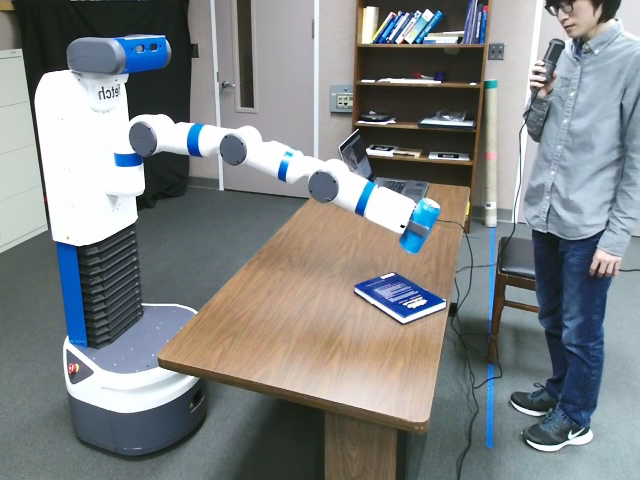}
  }
  \subfloat[][]
  {
    \includegraphics[width=0.32\linewidth]{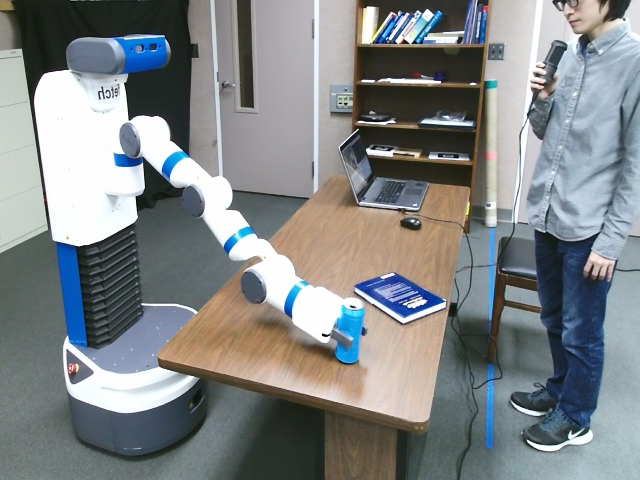}
  }
  \caption{The Fetch robot is moving a soda can on a table based on NLP instructions. Initially, the user gives the ``pick and place" command. However, when the robot gets closer to the book, the person says \emph{``don't put it there''} (i.e. negation) and the robot uses our dynamic constraint mapping functions and optimization-based planning to avoid the book. Our approach can generate appropriate motion plans for such attributes.}
  \label{fig:stop}
\end{figure}
At a high level, natural language instructions can be decomposed into task description and attributes. Task descriptions are usually verbs or noun phrases that describe the underlying task performed by a robot. The attributes include various adjectives, adverbs, or prepositional phrases that are used to specify additional conditions the robot must (or must not) satisfy. For example, these conditions may specify some information related to the movement speed, the orientation, the physical space characteristics, or the  distances. Therefore, it is important to design motion planners that consider these robotic task descriptions and robot motion constraints.

\noindent {\bf Main Results:}  We present an algorithm for generating parameterized constraints for optimization-based motion planning from complex, attribute-based natural language instructions. We use  {\em Dynamic Grounding Graphs} (DGG) to parse and interpret the commands and to generate the constraints. Our formulation includes the latent parameters in the grounding process, allowing us to model many continuous variables in our grounding graph. Furthermore, we present a new dynamic constraint mapping that takes DGG as the input and computes different constraints and parameters for the motion planner. The appropriate motion parameters are speed, orientation, position, smoothness, repulsion, and avoidance. The final trajectory of the robot is computed using a constraint optimization solver. Overall, our approach can automatically handle complex natural language instructions corresponding to spatial and temporal adjectives, adverbs, superlative and comparative degrees, negations, etc. Compared to prior techniques, our overall approach offers the following benefits:

\begin{itemize}
\item The inclusion of latent parameters in the grounding graph allows us to model continuous variables that are used by our mapping algorithm. Our formulation computes the dynamic grounding graph based on conditional random fields.
\item We present a novel dynamic constraint mapping used to compute different parametric constraints  for  optimization-based motion planning.
\item Our grounding graphs can handle more complex, attribute-based natural language instructions, and our mapping algorithm uses appropriate cost functions as parameters over the continuous space. Compared to prior methods, our approach is much faster and better able to handle more complex and attribute-based natural language instructions.
\end{itemize}

We highlight the performance of our algorithms in a simulated environment and on a 7-DOF Fetch robot operating next to a human. Our approach can handle a rich set of natural language commands and can generate appropriate paths. These include complex commands such as  picking (e.g., \emph{``pick up a red object near you"}), correcting the motion  (e.g., \emph{``don't pick up that one"}), and negation (e.g., \emph{``don't put it on the book"}).

\section{Related Work}
Most algorithms used to map natural language instruction to robot actions tend to separate the problem into two parts: parsing and motion planning computation. In this section, we give a brief overview of prior work in these areas.

\nocite{dragan2013legibility}

\subsection{Natural Language Processing}

Duvallet et al.~\cite{duvallet2016inferring} use a probabilistic graphical learning model called Generalized Grounding Graphs (G$^3$) on a ground vehicle for a navigation problem given natural language commands. Branavan et al.~\cite{branavan2009reinforcement,branavan2012learning} use reinforcement learning to learn a mapping from natural language instructions and then apply it to sequences of executable actions. Matuszek et al.~\cite{matuszek2010following} use a statistical machine translation model to map natural language instructions to path description language, which allows a robot to navigate while following directions. Duvallet et al.~\cite{duvallet2013imitation} use imitation learning to train the model through demonstrations of humans following directions. Paul et al.~\cite{paul2016efficient} propose the Adaptive Distributed Correspondence Graph (ADCG). Arkin et al.~\cite{arkin2015towards} further extend DCG, proposing the Hierarchical Distributed Correspondence Graph (HDCG), which  defines constraints as discrete inequalities and grounds word phrases to corresponding inequalities. Chung et al.~\cite{chung2015performance} use HDCG on ground vehicles to implement navigation commands and demonstrate performance improvements over G$^3$ in terms of running time, factor evaluations, and correctness. Oh et al.~\cite{oh2016integrated} integrate HDCG with their navigating robot system, measuring performance in terms of completion rates and comparing them to human behaviors. Scalise et al.~\cite{scalise2018natural} collected a corpus of natural language instructions from online crowdsourcing that specify objects of interest for \emph{``picking up''} command. The dataset could be used as a training dataset in our method.

\subsection{Robot Motion Planning in Dynamic Environments}

Many replanning algorithms have been suggested to generate collision-free motion plans in dynamic environments. Fox et al.~\cite{fox1997dynamic} propose the dynamic window approach to compute optimal velocity in a short time window. Optimization-based motion planners~\cite{STOMP:2011,zucker2013chomp,GPUITOMP} solve a constrained optimization problem to generate smooth and collision-free robot paths. We present an automatic scheme that generates the motion planning problem from NLP instructions.

\begin{figure}[t]
  \centering
  \includegraphics[width=\linewidth]{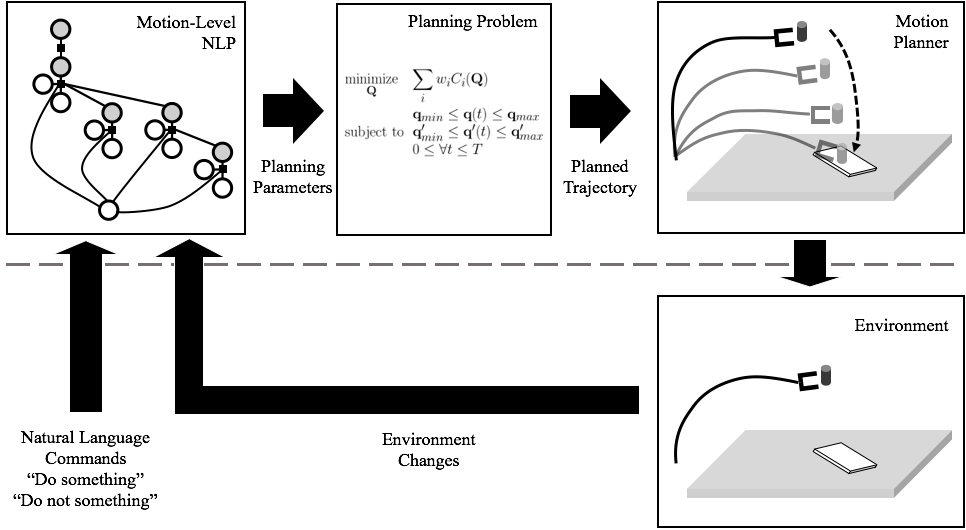}
  \caption{
  The overall pipeline of our approach highlighting the NLP parsing module and the motion planner. Above the dashed line (from left to right): Dynamic Grounding Graphs (DGG) with latent parameters that are used to parse and interpret the natural language commands, generation of optimization-based planning formulation with appropriate constraints and parameters using our mapping algorithm. We highlight the high-level interface below the dashed line. As the environment changes or new natural language instructions are given, our approach dynamically changes the specification of the constraints for the optimization-based motion planner and generates the new motion plans.
  }
  \label{fig:pipeline}
\end{figure}

There is some work on integrating optimization-based motion planning with NLP in 2D workspaces. Silver et al.~\cite{silver2013learning} develop an algorithm for learning navigation cost functions from demonstrations. Howard et al.~\cite{howard2014natural} use a probabilistic graphical model to generate motion planning constraints for a 2D navigation problem. Compared to these methods, our approach can handle 3D workspaces and high-dimensional configuration spaces to generate robot motions corresponding to complex NLP instructions. Other techniques focus on efficiency in human-robot collaborative tasks. Markov Decision Processes (MDP) are widely used to compute the best robot action policies~\cite{nikolaidis2013human,koppula2016anticipatory}. These techniques are complementary to our approach.

\section{Dynamic Grounding Graphs}
\label{sec:dgg}
Fig.~\ref{fig:pipeline} shows the basic pipeline of our approach. When natural language commands are given as input, the NLP module (upper left) creates an optimization problem for a motion planning module (upper middle). The robot motion trajectory is then computed from the motion planning module (upper right). As the planned trajectory is executed (bottom right), the result is fed back to the NLP module. In this section, we present the algorithms used in the NLP module.

\begin{figure}[t]
  \centering
  \subfloat
  {
    \includegraphics[width=0.28\linewidth]{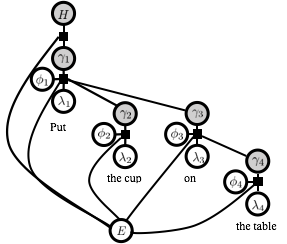}
  }
  \subfloat
  {
    \includegraphics[width=0.31\linewidth]{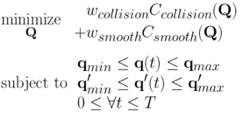}
  }
  \subfloat
  {
    \includegraphics[width=0.31\linewidth]{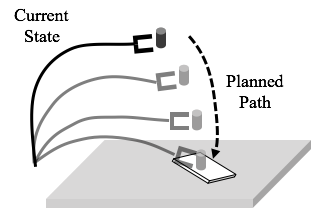}
  }
  \addtocounter{subfigure}{-1}
  \addtocounter{subfigure}{-1}
  \addtocounter{subfigure}{-1}
  \\
  \subfloat[][]
  {
    \includegraphics[width=0.28\linewidth]{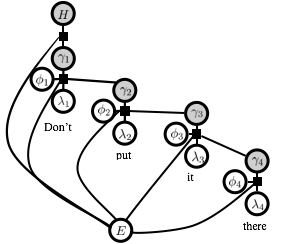}
  }
  \subfloat[][]
  {
    \includegraphics[width=0.31\linewidth]{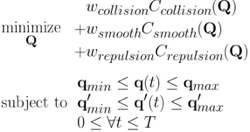}
  }
  \subfloat[][]
  {
    \includegraphics[width=0.31\linewidth]{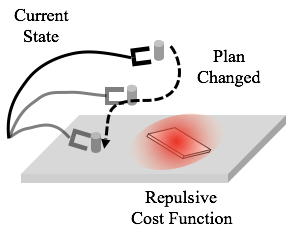}
  }
  \caption{{\bf  Factor graphs for different commands:} In the environment in the right-hand column, there is a table with a thin rectangular object on it. A robot arm is moving a cup onto the table, but we want it to avoid moving over the book when given NLP instructions.
  (a) The  command \emph{``Put the cup on the table"} is given and the factor graph is constructed (left).
  Appropriate cost functions for the task are assigned to the motion planning algorithm (middle) and used to compute the robot motion  (right).
  (b) As the robot gets close to the book, another command \emph{``Don't put it there"} is given with a new factor graph and cost functions.
 }
  \label{fig:nlp_overview}
\end{figure}

We extend the ideas of the Generalized Grounding Graphs (G$^3$) model and the Distributed Correspondence Graph (DCG) model~\cite{howard2014natural} by including the latent variables in the grounding graph and using them to compute the constraints for motion planning. The input to our algorithm is the natural language instruction. We do not account for any errors due to voice recognition. From a natural language command input, we construct a factor graph, as shown in Fig.~\ref{fig:nlp_overview}(a), which is based on the parsing of the command. For each node of the parse tree, we generate three types of nodes: word phrase node $\lambda$, grounding node $\gamma$, and correspondence node $\phi$.

The input sentence $\Lambda$ is parsed using the NLTK library~\cite{bird2006nltk}. The word phrase of each node in the parse tree is denoted as $\lambda_i$ for $i=1,2,\cdots$. Children of $\lambda_i$ are $\lambda_{i1}$, $\cdots$, $\lambda_{im}$. The root node of the parse tree is $\lambda_1$. For example, in Fig.~\ref{fig:nlp_overview}(a), the input sentence is \emph{``Put the cup on the table.''} The parse tree has the root word phrase $\lambda_1=$\emph{``Put''}. Its noun $\lambda_2=$\emph{``the cup"} and the preposition $\lambda_3=${``on,''} which are the children nodes of the root node. The noun phrase $\lambda_4=$\emph{``the table''} is the child node of $\lambda_3$. Similarly, in Fig.~\ref{fig:nlp_overview}(b), the command \emph{``Don't put it there''} is decomposed into 4 noun phrase nodes. The word phrase $\lambda_1=$\emph{``Don't''} is a negation of the verb and its child node is $\lambda_2=$\emph{``put.''} $\lambda_3=$\emph{``it''} and $\lambda_4=$\emph{``there''} are the children nodes of $\lambda_2$. Note that this parse tree is different from the parse tree in Fig.~\ref{fig:nlp_overview}(a).

Our goal is to compute a mapping from a natural language sentence $\Lambda$ to the cost function parameters $H$, given the robotic environment $E$ where the robot is operating. $E$ is a representation of the environment, which is composed of obstacle positions, orientations, and the robot's configuration. Feature vectors are constructed in the factor graph from the description of the environment. $H$ is a real-valued vector that contains all cost function parameters used in the optimization-based motion planner. It also includes the weights of different types of cost functions used in the optimization formulation. For example, the end-effector position cost function (Eq. (\ref{eq:target_cost})) requires the 3D coordinates of the target position as parameters. The repulsion cost function (Eq. (\ref{eq:repulsion_cost})) requires the repulsion source position and the constant from the exponential function.

We first compute the groundings $\gamma_i$ of each word phrase $\lambda_i$. The grounding of each word phrase is the mapping from the word phrase to its meaning in the real world. Groundings can be objects, locations, motions, tasks, or constraints. In our model, the grounding $\gamma_i$ depends on its work phrase $\lambda_i$ and its children grounding nodes $\gamma_{i1}$, $\cdots$, $\gamma_{im}$, where the tree structure of the grounding nodes follows the parse tree. Correspondence node $\phi_i$ indicates the correct matching between the word phrase $\lambda_i$ and the grounding $\gamma_i$. It is a binary variable; $\phi_i$ is $true$ if the word phrase and the grounding match and $false$ if they do not.

\subsection{Latent Parameters}

A key novel component of our approach is the inclusion of latent variables in the grounding graph. Our primary goal is to compute the best cost function parameters $H$ to be used directly for optimization-based motion planning. We denote $H \in \mathbb{R}^h$, a real vector of size $h$, as a collection of cost function parameters. In this case, the size $h$ and the number of cost function parameters depend on the types of cost functions that are used. \footnote{In this paper, we set a maximum $h=22$ to fully specify the smoothness, the end-effector position, the end-effector orientation, the end-effector speed, and the repulsion cost functions. It is a sum of three terms: $5$ for weights, $16$ for positions and orientations, and $1$ for an exponential constant.} From the predicted groundings $\gamma_i$, the cost function parameters in the motion planning formulation (Fig.~\ref{fig:nlp_overview}(b)) are inferred through the latent variable $H$. $H$ contains all the cost function parameters (e.g., weights of cost functions, locations, and orientations).

In Fig.~\ref{fig:nlp_overview}(b), the resulting constraint-based motion planning problems are shown. We use the collision avoidance cost function as the default smoothness cost function and the target location cost function, though weights can vary. The target location, whose 3D coordinates are the cost function parameters, is set on the surface of the table. The cost function parameter node $H$ contains the weights of the parameters and the 3D coordinates of the target location. In the bottom of Fig.~\ref{fig:nlp_overview}(b), where a new \emph{``Don't"} command is given, a repulsion cost function is added. Thus, the cost function weight and the location of the repulsion source (below the robot’s end-effector position) are added to $H$.

\subsection{Probabilistic Model}
\label{subsec:probabilistic_model}
We present a new probabilistic model to compute $H$, $\Lambda$, and $E$. We pose the problem of finding the best cost parameters as an optimization problem:
\begin{equation}
\begin{aligned}
& \underset{\mathbf{H}}{\text{maximize}}
& & p(H | \Lambda, E).
\end{aligned}
\end{equation}
However, modeling the probability function without decomposing the variables and some assumptions about independence is difficult due to the high-dimensionality of $H$, $\Lambda$, and $E$ and the dependencies between them. To simplify the problem, the natural language sentence is decomposed into $n$ word phrases based on a parse tree, i.e.
\begin{equation}
p(H | \Lambda, E) = p(H | \lambda_1, \cdots, \lambda_n, E).
\end{equation}
Like G$^3$, we introduce the intermediate groundings $\gamma_i$ of word phrases $\lambda_i$ and correspondence variables $\phi_i$. The correspondence variables $\phi_i$ are binary random variables.
The value $1$ indicates that the word phrase $\lambda_i$ correctly corresponds to the grounding $\gamma_i$. $0$ means an incorrect correspondence.

We assume the conditional independence of the probabilities so that we can construct a factor graph (see Fig.~\ref{fig:nlp_overview}(a)). With the independence assumptions, a single factor is connected to a word phrase node and its children grounding nodes, which contain information about the sub-components. These independence assumptions simplify the problem and make it solvable by efficiently taking advantage of the tree structure of the probabilistic graphical learning model. Formally, the root grounding node $\gamma_1$ contains all the information about a robot's motion. The factor that connects $\gamma_1$ and $H$ implies that, from the root grounding node, the cost function parameters $H$ are optimized without any consideration of other nodes. Other factors connect $\gamma_i$, $\phi_i$, $\lambda_i$, children grounding nodes $\gamma_{ij}$ and the environment $E$, where the parent-child relationship is based on a parse tree constructed from the natural language sentence. This graphical representation corresponds to the following equation:
\begin{align}
p(H | \lambda_1, \cdots, \lambda_n, E)  
= p(H | \gamma_1, E) \prod_i p(\gamma_i | \lambda_i, \phi_i, \gamma_{i1}, \cdots, \gamma_{im}, E).
\end{align}
For the root factor connecting $H$, $\gamma_1$ and $E$, we formulate the continuous domain of $H$. We compute the Gaussian Mixture Model (GMM) on the probability distribution $p(H | \gamma_1, E)$ and model our probability with non-root factors as follows:
\begin{align}
 & p(\gamma_i | \lambda_i, \phi_i, \gamma_{i1}, \cdots, \gamma_{im}, E) \nonumber \\
=& \frac{1}{Z} \psi_i(\gamma_i, \lambda_i, \phi_i, \gamma_{i1}, \cdots, \gamma_{im}, E) \nonumber \\
=& \frac{1}{Z} \exp(- \theta_i^T f(\gamma_i, \lambda_i, \phi_i, \gamma_{i1}, \cdots, \gamma_{im}, E)), \label{eq:one_sample}
\end{align}
where $Z$ is the normalization factor, and $\theta_i$ and $f$ are the log-linearization of the feature function. The function $f$ generates a feature vector given a grounding $\gamma_i$, a word phrase $\lambda_i$, a correspondence $\phi_i$, children groundings $\gamma_{ij}$, and the environment $E$. The information from the robotic environment is used in the feature function $f$ and in the log-linearized feature function $f$. The attributes of objects in the robotic world such as shapes and colors are encoded as multidimensional binary vectors, which indicate whether the object has a given attribute.

The probability distribution of the latent variable $H$ is modeled with $m$ pairs of Gaussian distribution parameters $\mu_i$ and $\sigma_i$ with weights $\omega_i$, as follows:
\begin{align}
p(H | \gamma_1, E) \sim \sum_{i=1}^{m} \omega_i \mathcal{N}(\mu_i, \sigma_i^2).
\label{eq:gmm}
\end{align}

\noindent \emph{\bf Word phrases.}
The feature vector includes binary-valued vectors for the word and phrase occurrences, and Part of Speech (PoS) tags. There is a list of words that could be encountered in the training dataset such as {\em \{put, pick, cup, up, there, $\cdots$ \}}. If the word phrase contains the word {\em ``put,''} then the occurrence vector at the first index is set to 1 and the others are set to 0. If the word phrase is {\em ``pick up,''} then the occurrence vector values at the second, while the fourth is set to 1 and others are set to 0. This list also includes real-valued word similarities between the word and the pre-defined seed words. The seed words are the pre-defined words that the users expect to encounter in the natural language instructions. We used Glove word2vec~\cite{pennington2014glove} to measure cosine-similarity (i.e. the inner product of two vectors divided by the lengths of the vectors) between the words. The measurement indicates that the words are similar if the similarity metric  value is near $1$, that they have opposite meanings if the similarity metric is near -1, and that they have a weak relationship if it is near 0. This provides more flexibility to our model, especially when it encounters new words that are not trained during the training phase.

\noindent \emph{\bf Robot states.}
From the robot state, we collect the robot joint angles, the velocities, the end-effector position, the end-effector velocity, etc. This information can affect the cost function parameters even while processing the same natural language commands. For example, if the robot is too close to a human under the current configuration, then the cost function for end-effector speed $C_{speed}$ or smoothness $C_{smoothness}$ will be adjusted so that the robot does not collide with the human. We also store information about the objects that are close to the robot. This information includes object type, position, orientation, shape, dimension, etc.

\subsection{Factor Graph using Conditional Random Fields}
We represent our dynamic grounding graph as a factor graph. We build a factor graph based on the probabilistic model described in Section~\ref{subsec:probabilistic_model} and use that for training and for inferring the meaning of given commands. In particular, we use Conditional Random Fields (CRF)~\cite{sutton2012introduction} as a learning model for factor graphs because CRFs are a good fit for applying machine learning to our probabilistic graph model with conditional probabilities.

During the training step of CRF, we solve the optimization problem of maximizing the probability of the samples in the training dataset over the feature coefficients $\theta_i$ and the GMM parameters $\omega_i$, $\mu_i$ and $\sigma_i$ for every parse tree structure. By multiplying Eq. (\ref{eq:one_sample}-\ref{eq:gmm}) for all training samples, the optimization problem becomes
\begin{align}
\underset{\substack{\theta_1, \cdots, \theta_n, \\ \omega_1, \cdots, \omega_m, \\ \mu_1, \cdots, \mu_m, \\ \sigma_1, \cdots, \sigma_m }}{\text{maximize}} & \prod_k p(H^{(k)} | \gamma_1^{(k)}, E^{(k)}) \label{eq:training_gmm} \\
& \prod_i \frac{1}{Z^{(k)}} \exp(\theta_i^T f(\gamma_i^{(k)}, \lambda_i^{(k)}, \phi_i^{(k)}, \gamma_{i1}^{(k)}, \cdots, \gamma_{im}^{(k)}, E^{(k)})), \label{eq:training_crf}
\end{align}
where superscripts $(k) = 1 \cdots D$ mean the indices of the training samples. The joint optimization problem Eq. (\ref{eq:training_gmm}-\ref{eq:training_crf}) of the GMM and the CRF is a hard problem. So, we separate the problem into two and solve each one separately to maximize the objective. To solve Eq. (\ref{eq:training_gmm}), the training samples of continuous variable $H$ is collected under the same conditional variable $\gamma_{1}^{(k)}$. Then, we solve the problem with the collection of $H$'s via Expectation Maximization (EM) method. Eq. (\ref{eq:training_crf}) is a tree-structured CRF problem.

At the inference step, we used the trained CRF factor graph models to find the best groundings $\Gamma$ and the cost function parameters $H$ by solving the CRF maximization problem
\begin{align}
\underset{H, \gamma_1, \cdots, \gamma_n}{\text{maximize}} & p(H | \gamma_1, E) 
\prod_i \frac{1}{Z} \exp(\theta_i^T f(\gamma_i, \lambda_i, \phi_i, \gamma_{i1}, \cdots, \gamma_{im}, E)) .
\label{eq:inference}
\end{align}
When the nodes $H$, $\gamma_1, \cdots, \gamma_n$ are optimized, they create a tree structure in the factor graph, meaning that we can solve the optimization problem efficiently using dynamic programming. Each factor depends on its parent and children varying variables and other fixed variables connected to it. This implies that we can solve the sub-problems in a bottom-up manner and combine the results to solve the bigger problem corresponding to the root node.

\section{Dynamic Constraint Mapping With NLP Input}
\label{sec:planning}

We use an optimization-based algorithm~\cite{Park:2012:ICAPS} to solve the cost minimization problem. The function and constraints of this cost minimization problem come from DGG, as explained in Sec.~\ref{sec:dgg}. In this section, we present our mapping algorithm, Dynamic Constraint Mapping, which maps the word phrase groundings to proper cost function parameters corresponding to natural language instructions.

\subsection{Robot Configurations and Motion Plans}
\label{subsec:planning}

We denote a single configuration of the robot as a vector $\mathbf{q}$, which consists of joint-angles or other degrees-of-freedom. A configuration at time $t$, where $t \in \mathbb{R}$, is denoted as $\mathbf{q}(t)$. We assume $\mathbf{q}(t)$ is twice differentiable, and its derivatives are denoted as $\mathbf{q}'(t)$ and $\mathbf{q}''(t)$. The $n$-dimensional space of configuration $\mathbf{q}$ is the configuration space $\mathcal{C}$. We represent bounding boxes of each link of the robot as $B_i$. The bounding boxes at a configuration $\mathbf{q}$ are denoted as $B_i(\mathbf{q})$.

For a planning task with a given start configuration $\mathbf{q}_0$ and derivative $\mathbf{q}'_0$, the robot's trajectory is represented by a matrix $\mathbf{Q}$, whose elements correspond to the waypoints~\cite{Zucker:IJRR:2012,STOMP:2011,zucker2013chomp,Park:2012:ICAPS}:
\begin{equation}
\mathbf{Q} = \begin{bmatrix}
    \mathbf{q}_0 & \mathbf{q}_1 & & \mathbf{q}_{n-1} & \mathbf{q}_n \\
    \mathbf{q}'_0 & \mathbf{q}'_1 & \cdots & \mathbf{q}'_{n-1} & \mathbf{q}'_n \\
    t_0 = 0 & t_1 & & t_{n-1} & t_n = T
\end{bmatrix}.
\end{equation}
The robot trajectory passes through $n+1$ waypoints $q_{0}, \cdots, q_{n}$, which will be optimized by an objective function under constraints in the motion planning formulation. Robot configuration at time $t$ is interpolated from two waypoints. Formally, for $j$ such that $t_j \leq t \leq t_{j+1}$, the configuration $\mathbf{q}(t)$ and derivative $\mathbf{q}'(t)$ are cubically interpolated using $\mathbf{q}_j$, $\mathbf{q}'_j$, $\mathbf{q}_{j+1}$, and $\mathbf{q}'_{j+1}$.

The $i$-th cost functions of the motion planner are $C_i(\mathbf{Q})$. Our motion planner solves an optimization problem with non-linear cost functions and linear joint limit constraints to generate robot trajectories for time interval $[0, T]$,
\begin{equation}
\begin{aligned}
& \underset{\mathbf{Q}}{\text{minimize}}
& & \sum_i w_i C_i(\mathbf{Q}) \\ \label{optimize}
& \text{subject to}
& &
\begin{array}{l}
  \mathbf{q}_{min} \leq \mathbf{q}(t) \leq \mathbf{q}_{max}, \\
  \mathbf{q}'_{min} \leq \mathbf{q}'(t) \leq \mathbf{q}'_{max}
\end{array}
\, \, 0 \leq \forall t \leq T.
\end{aligned}
\vspace{-6pt}
\end{equation}
In the optimization formulation, $C_i$ is the $i$-th cost function and $w_i$ is the weight of the cost function.

\subsection{Cost Functions}

The overall optimization formulation is given in Eq. (\ref{optimize}). To formulate the constraints, we use the following cost functions, which are designed to account for various attributes in the NLP instructions.
In our formulation, we use many types of cost functions such as collision avoidance, robot smoothness, robot end-effector speed, target positions, and target orientations. These cost functions are used to handle many attributes of natural language instructions. Each cost function has its weight and may also have other cost function parameters, if necessary. For example, the robot end-effector speed cost function has parameters corresponding to the direction and the magnitude of the speed, which impose a constraint on the final computed trajectory. If the weight of the end-effector speed cost function is higher than the others, then it contributes more to the overall objective function in the optimization formulation. If the weight is low, then the end-effector speed cost will be compromised and has a lesser impact on the path planner.

The cost functions $C_i$ and the latent parameter $H$ are closely related. $H$ is a collection of parameters that describe all types of $C_i$ and the weights $w_i$. The cost function parameters of $C_i$ and the weights $w_i$ are all real-valued. Those real values are appended to construct the real-valued vector $H$.

\subsection{Parameterized Constraints}

To handle various attributes, we use the following parameterized constraints in our optimization formulation.

\noindent \emph{\bf Collision avoidance:} By default, the robot should always avoid obstacles.
\begin{align}
C_{collision}(\mathbf{Q}) = \int_0^T \sum_i \sum_j \textit{dist}(B_i(t), O_j)^2 dt,
\end{align}
where $\textit{dist}(B_i(t), O_j)$ is the penetration depth between a robot bounding box $B_i(t)$ and an obstacle $O_j$.

\noindent \emph{\bf Smoothness:} We penalize the magnitude of a robot's joint angle speed to make the  trajectory smooth. This corresponds to the integral of the first derivative of joint angles over the trajectory duration, as follows:
\begin{align}
C_{smoothness}(\mathbf{Q}) = \int_0^T \sum_i \mathbf{q}'(t)_i^2 dt.
\end{align}
This function is useful when we need to control the speed of the robot. When the robot should operate at a low speed (e.g. when a human is too close), or we don't want abrupt movements (e.g., for human safety), the smoothness cost can have high weights so that the robot moves slowly without jerky motions.

\noindent \emph{\bf End-effector position:} A user usually specifies the robot's target position to make sure that the robot reaches its goal. This cost function penalizes the squared distance between the robot's end-effector position and the target position over the trajectory duration as
\begin{align}
C_{position}(\mathbf{Q}) = \int_0^T || \mathbf{p}_{ee}(t) - \mathbf{p}_{target} ||^2 dt, \label{eq:target_cost},
\end{align}
where $\mathbf{p}_{ee}(t)$ is the robot end-effector position at time $t$ and $\mathbf{p}_{target}$ is the target position.
The target position $\mathbf{p}_{target}$ is considered as a cost function parameter. In the mapping algorithm, a position grounding node encodes the target position parameter. This parameter can be a 3D position or the current object position in the environment. Typically, the target position is specified by an object name in the sentence, such as \emph{``pick up the cup''} or \emph{``move to the box.''} In these cases, the grounding nodes for \emph{``the cup"} and \emph{``the box''} are interpreted as the current 3D coordinates of the target positions, which are the parameters of this cost function.

\noindent \emph{\bf End-effector orientation:} Robotic manipulation tasks are sometimes constrained by the end-effector orientation. This cost function penalizes the squared angular differences between the end-effector orientation and the target orientation over the trajectory duration.
\begin{align}
C_{orientation}(\mathbf{Q}) = \int_0^T \textit{angledist}(\mathbf{q}_{ee}(t), \mathbf{q}_{target})^2 dt \\
C_{upvector}(\mathbf{Q}) = \int_0^T \textit{angledist}(\mathbf{n}_{up}(t), \mathbf{n}_{target})^2 dt,
\end{align}
where $\mathbf{q}_{ee}(t)$ is the quaternion representation of the robot end-effector's orientation at time $t$, $\mathbf{q}_{target}$ is the end-effector orientation that we want the robot to maintain, $\mathbf{n}_{up}$ is the normal up-vector of the robot's end-effector, and $\mathbf{n}_{target}$ is the target up-vector. As with the \emph{end-effector position} cost, the target orientation $\mathbf{q}_{target}$ is the cost function parameter. The target orientation usually depends on the object the robot picked up. For example, when the robot is doing a peg-hole insertion task under the command \emph{``insert that into the hole,''} the orientation of the robot's end-effector $\mathbf{q}_{ee}$ should be constrained near the hole. If the robot arm is holding a cup of water, it should be upright so it does not spill the water. In this case, $\mathbf{n}_{target}$ is set to $(0, 0, 1)$.

\noindent \emph{\bf End-effector speed:} This cost function penalizes the robot's end-effector speed and direction:
\begin{align}
C_{speed}(\mathbf{Q}) = \int_0^T || \mathbf{v}_{ee}(t) - \mathbf{v}_{target} ||^2 dt,
\end{align}
where $\mathbf{v}_{ee}(t)$ is the robot's end-effector speed at time $t$, and $\mathbf{v}_{target}$ is the target speed. The parameters of this cost function correspond to $\mathbf{v}_{target}$. In some cases, we must restrict the robot's end-effector velocity, e.g., if a user wants to pick up a cup filled with water and doesn't want to spill it. Spilling can be prevented by limiting the end-effector speed, making the robot move more slowly.

\noindent \emph{\bf Repulsion:} The repulsion functions are commonly represented as potential fields
\begin{align}
C_{repulsion}(\mathbf{Q}) = \int_0^T \exp \left( - c || \mathbf{p}_{ee}(t) - \mathbf{p}_{r} || \right) dt, \label{eq:repulsion_cost}
\end{align}
where $\mathbf{p}_{r}$ is the position to which we don't want the robot to move. The coefficient $c > 0$ suggests how much the cost is affected by $||\mathbf{p}_{ee}(t) - \mathbf{p}_{repulsive}||$, the distance between the end-effector position and the repulsion source. The cost function is maximized when the end-effector position is exactly at the repulsion source, and it decreases as the distance between the end-effector and the repulsion position increases.  For example, if the command is \emph{``Don't put the cup on the laptop,''} we can define a repulsion cost with the laptop position as the repulsion source. The cost function is inversely proportional to the distance between the end-effector and the laptop.

\section{Implementation and Results}

We have implemented our algorithm and evaluated its performance in a simulated environment and on a 7-DOF Fetch robot. All the timings are generated on a multi-core PC.with Intel i7-4790 8-core 3.60GHz CPU and a 16GB RAM. We use multiple cores for fast evaluation and parallel trajectory search to compute a good solution to the constrained optimization problem~\cite{Park:2012:ICAPS}.

\subsection{Training DGGs for Demonstrations}

We describe how the training dataset for our DGG model was generated. The training dataset for DGGs requires three components: a natural language sentence, a robotic environment, and the cost function parameters for optimization-based motion planners.

For each demonstration, we write tens of different sentences that specify the take goals the constraints for the motion plans with different nouns, pronouns, adjectives, verbs, adverbs, preposition, etc. For each sentence, we generate a random robotic environment and an initial state for the robot. In addition, the robot joint values and joint velocities are randomly set as initial states. We collect tens or hundreds of random robotic environments. For a natural language sentence, a random robotic environment, and a random initial state for the robot, the cost function parameters are assigned manually or synthesized from other examples. Crowdsourcing such as Amazon Mechanical Turk can be alternatively used to assign cost function parameters. Hundreds of multiple data samples are generated from generated data samples by switching the correspondence variable in the DGG model from $1$ (true) to $0$ (false) and changing the grounding variables to the wrong ones to match the false correspondence variable. The training dataset is created with up to $100,000$ samples in our experiments. When the cost function parameters are determined, the optimization-based motion planner is used to compute a feasible robot trajectory. In the optimization-based motion planning algorithm, there are some waypoints through which the robot trajectory should pass. For the robot's safety, we check if the robot trajectory with the given cost function parameters is in-collision and appropriately set a higher value of the coefficient of the collision cost and compute a new trajectory. This process is repeated until the trajectory is collision-free. The training step took up to an hour with $100,000$ training samples in our experimental settings, though the training time can vary depending on the complexity of tasks, environments, and natural language instructions.

\begin{figure}[t]
  \centering
  \subfloat[][]
  {
    \includegraphics[width=0.32\linewidth]{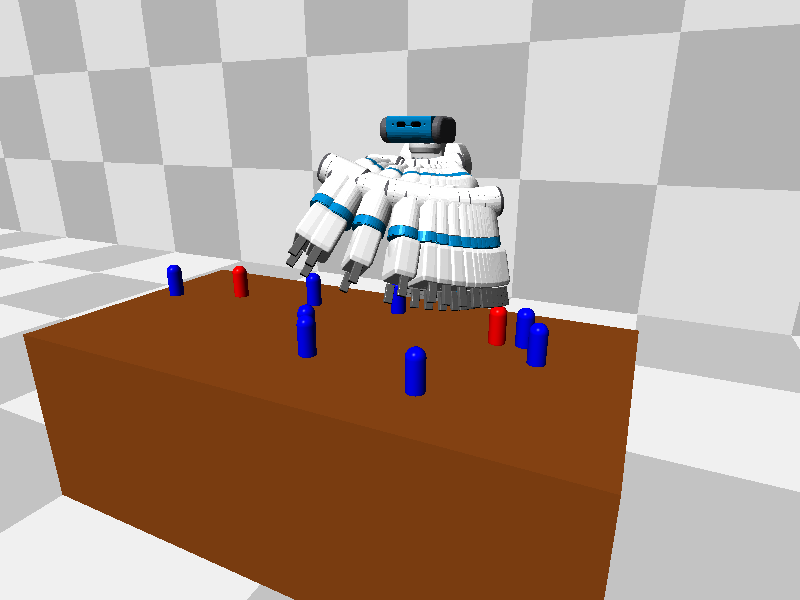}
  }
  \subfloat[][]
  {
    \includegraphics[width=0.32\linewidth]{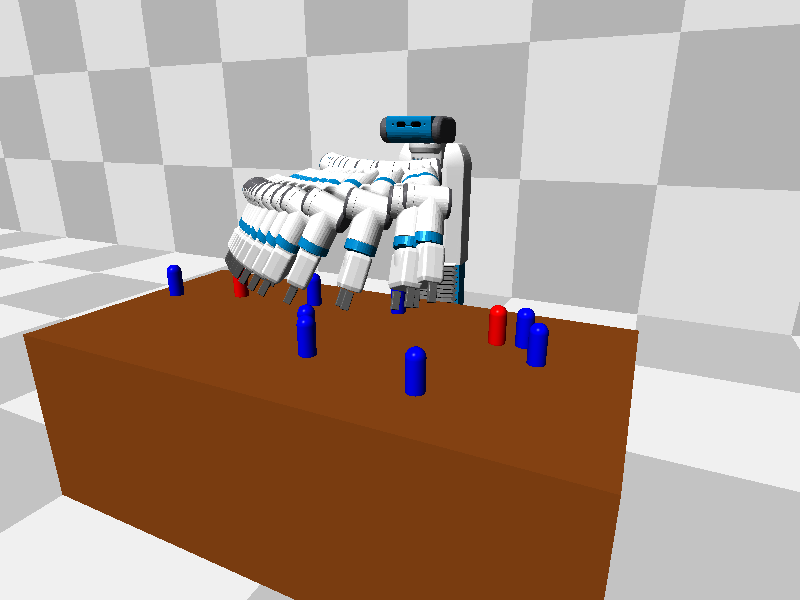}
  }
  \subfloat[][]
  {
    \includegraphics[width=0.32\linewidth]{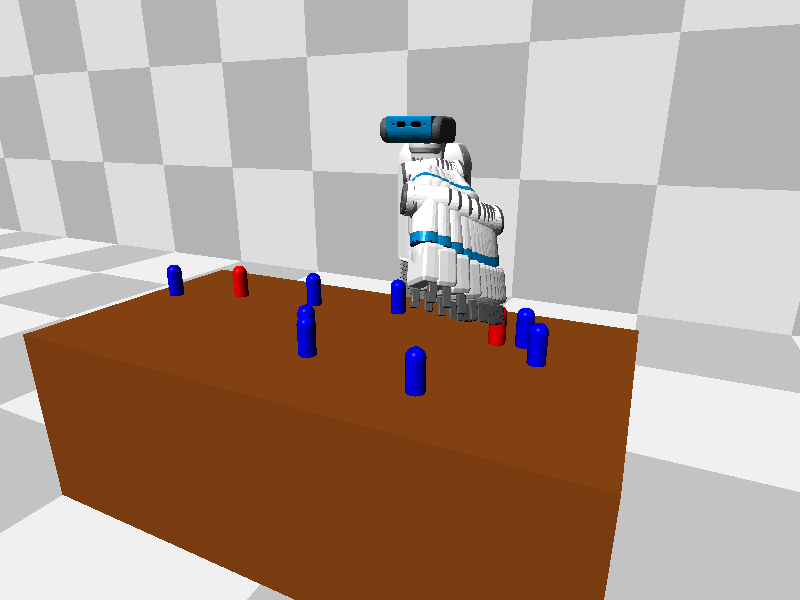}
  }
  \caption{The simulated Fetch robot arm reaches towards one of the two red objects.
  (a) When a command \emph{``pick up one of the red objects''} is issued, the robot moves to the red object on the right because of the DGG algorithm.
  (b) If the user doesn't want the robot to pick up the object on the right, he/she uses a command \emph{``don't pick up that one.''} Our DGG algorithm dynamically changes the cost function parameters. 
  (c) The robot approaches the object on the right and stops.
  }
  \label{fig:sim_1}
  \vspace*{-0.05in}
\end{figure}

\begin{figure}[t]
  \centering
  \subfloat[][]
  {
    \includegraphics[width=0.32\linewidth]{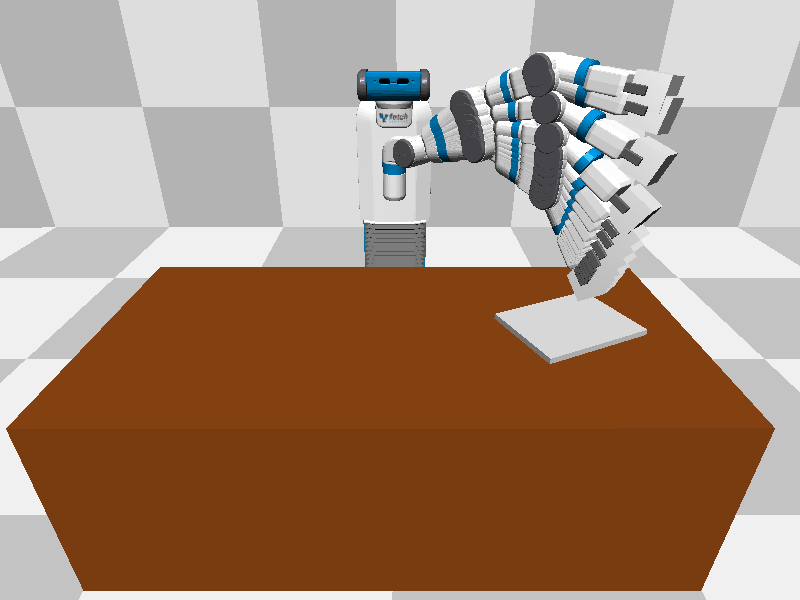}
  }
  \subfloat[][]
  {
    \includegraphics[width=0.32\linewidth]{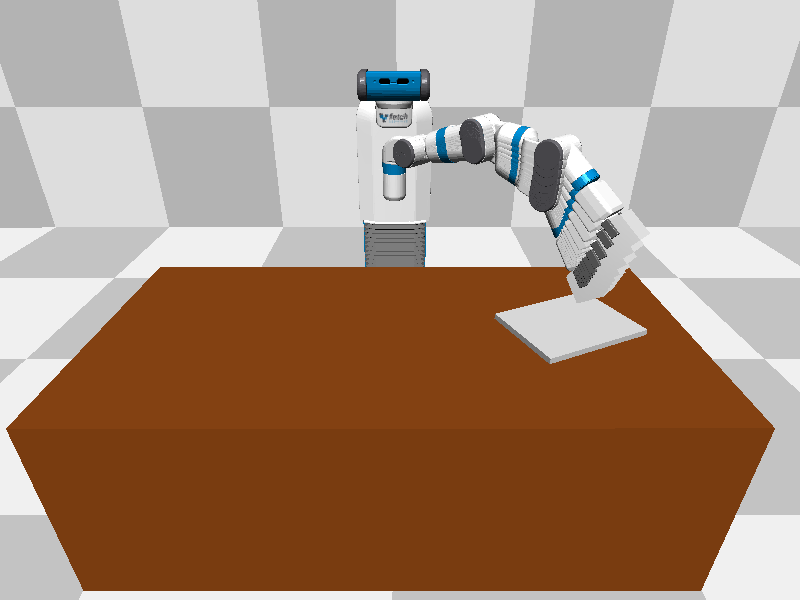}
  }
  \subfloat[][]
  {
    \includegraphics[width=0.32\linewidth]{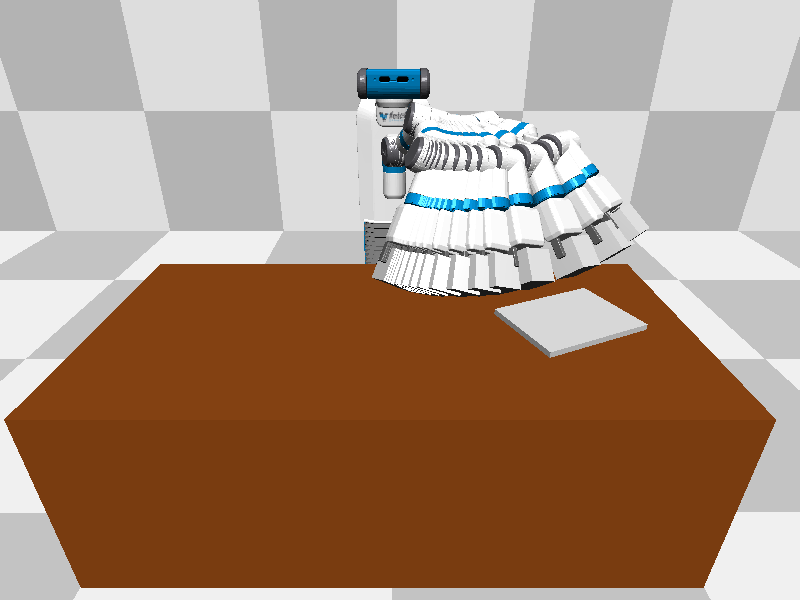}
  }
  \caption{In this simulated environment, the human instructs the robot to \emph{``put the cube on the table''} (a). As it approaches the laptop (b), the human uses a negation NLP command \emph{``don't put it there,''} so the robot places it at a different location (c).}
  \label{fig:sim_2}
\end{figure}

We use different training data for each scenario. For the scenarios shown in Fig.~\ref{fig:sim_1}, the initial pose of the robot in front of the table and the positions of the blue and red objects on the table are randomly set. For \textit{``Pick up''} commands, appropriate cost function parameters are computed so that the robot picks up a blue or red object depending on the given command. Similarly, in Fig.~\ref{fig:sim_2}, the position and orientation of the laptop is initialized randomly. Given the \textit{``Put''} command, we create an end-effector position cost function so that the robot places the object on the table; and a repulsive cost function to avoid the laptop position.

\subsection{Simulations and Real Robot Demonstrations}

\begin{figure}[ht]
  \centering
  \subfloat[][]
  {
    \includegraphics[width=0.32\linewidth]{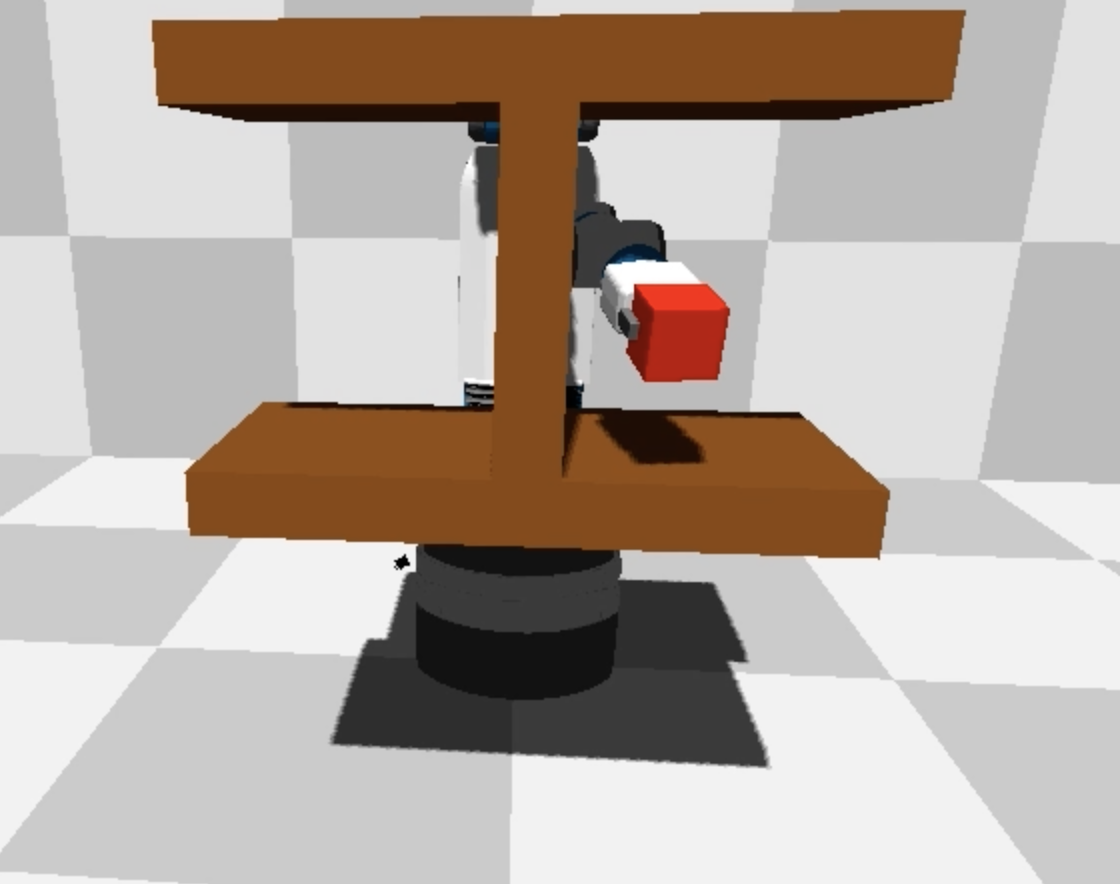}
  }
  \subfloat[][]
  {
    \includegraphics[width=0.32\linewidth]{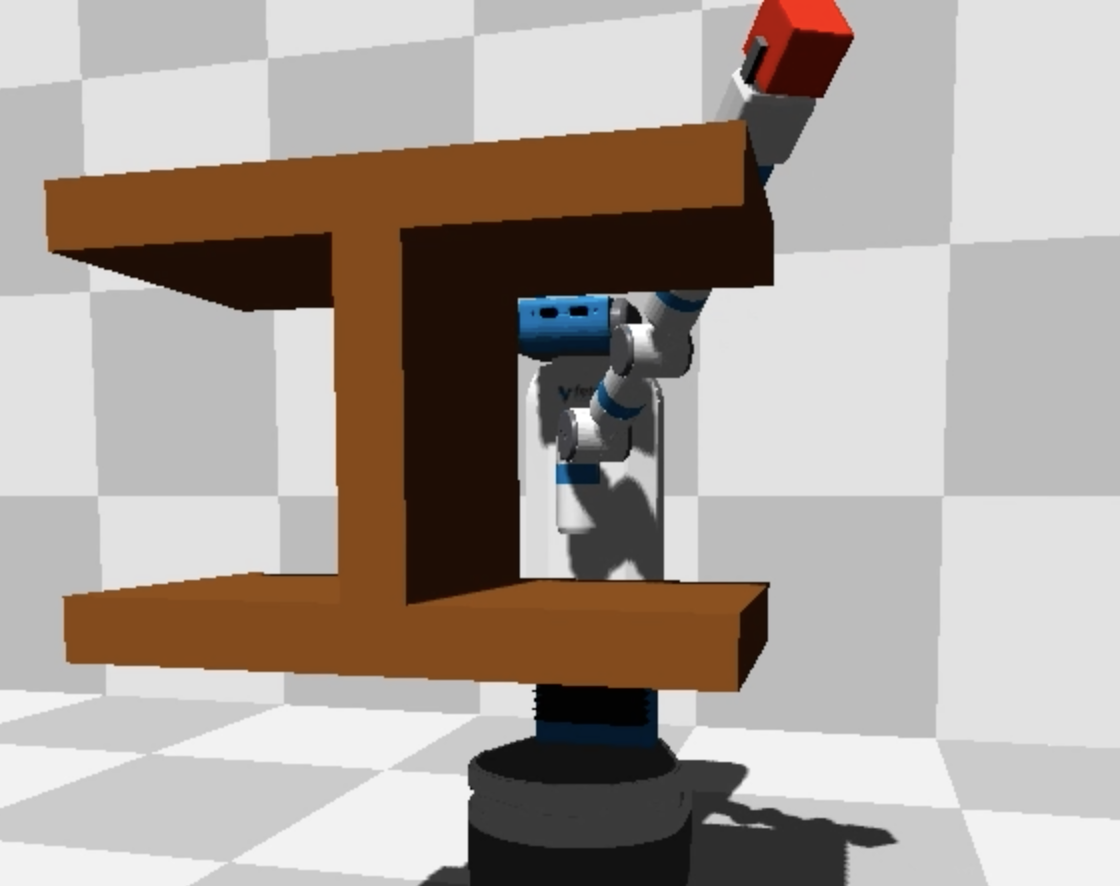}
  }
  \subfloat[][]
  {
    \includegraphics[width=0.32\linewidth]{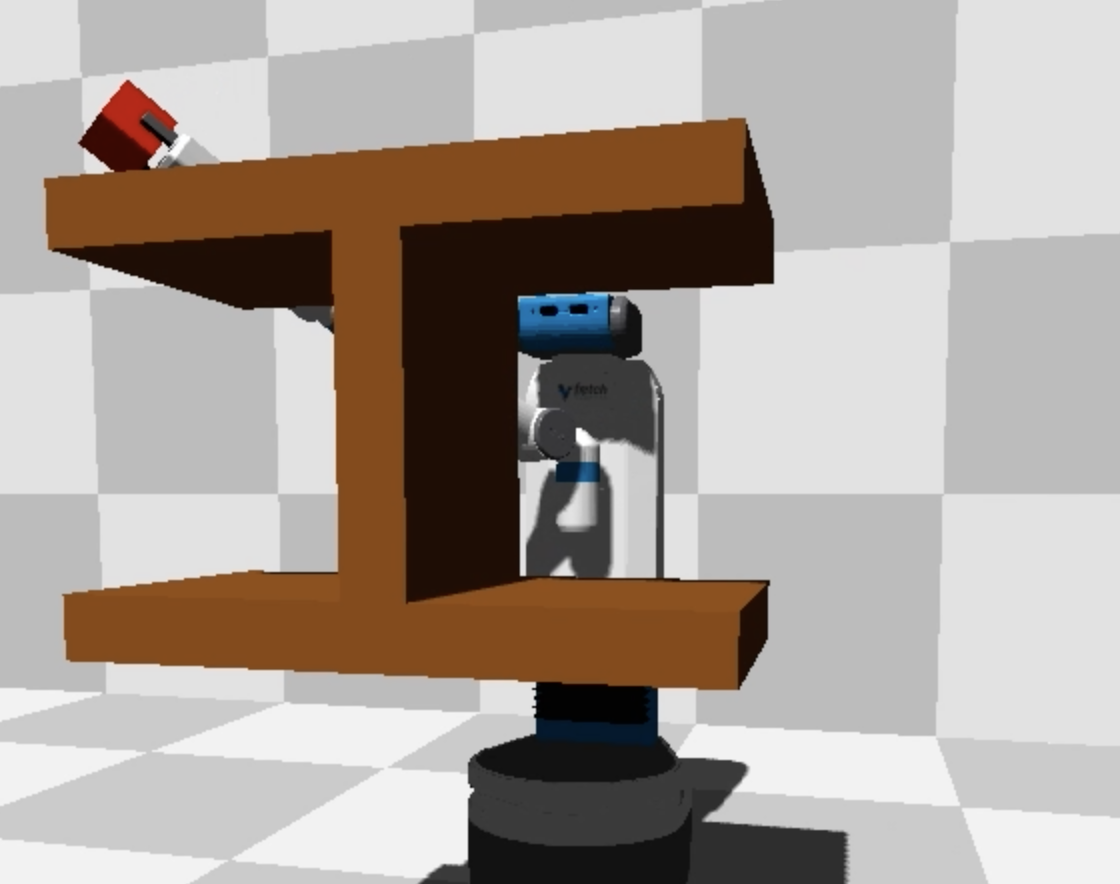}
  }
  \caption{A 7-DOF Fetch robot is operating in a simulated environment avoiding an obstacle.
  (a) In a traditional optimization-based motion planner, the planner gets stuck at a local minimum.
  (b),(c) Using natural language commands as guidance, the user guides the robot out of the minimum and towards the goal position.}
  \label{fig:sim_obstacle}
\end{figure}

We evaluate the performance on optimization problems that occur in complex environments composed of multiple objects. Based on the NLP commands, the robot decides to pick an appropriate object or is steered towards the goal position in a complex scene.  In particular, the user gives NLP commands such as \emph{``move right,''} \emph{``move up,''} \emph{``move left,''} or \emph{``move down''} to guide the robot. For each such command, we compute the appropriate cost functions. 

\begin{figure}[t]
  \centering
  \subfloat[][]
  {
    \includegraphics[width=0.32\linewidth]{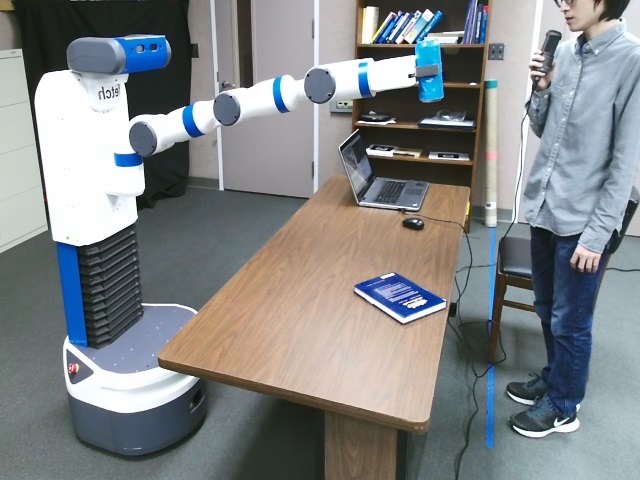}
  }
  \subfloat[][]
  {
    \includegraphics[width=0.32\linewidth]{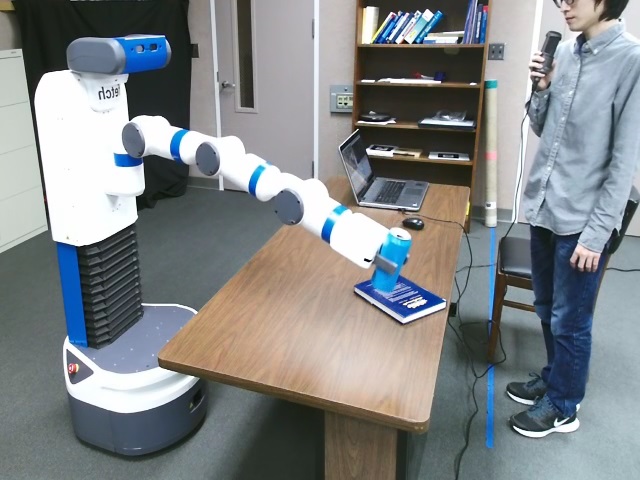}
  }
  \subfloat[][]
  {
    \includegraphics[width=0.32\linewidth]{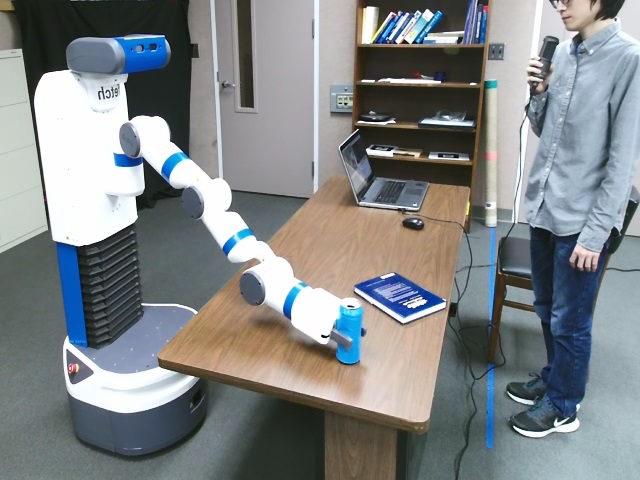}
  }
  \caption{The Fetch robot is taking real-time commands from the human and moves the soda can on the table.}
  \label{fig:dont}
\end{figure}

\begin{figure}[t]
  \centering
  \subfloat[][]
  {
    \includegraphics[width=0.32\linewidth]{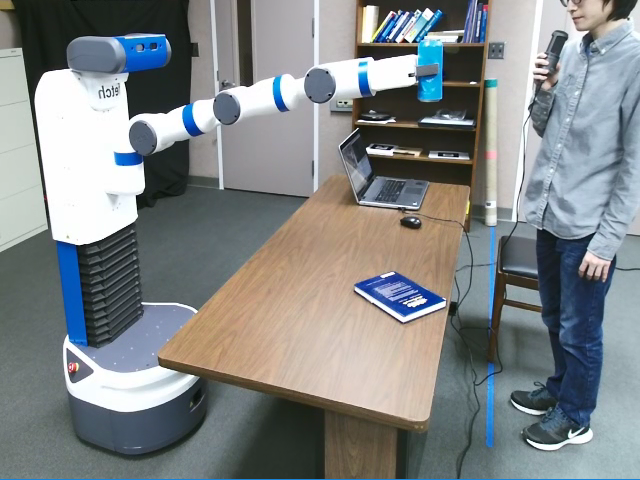}
  }
  \subfloat[][]
  {
    \includegraphics[width=0.32\linewidth]{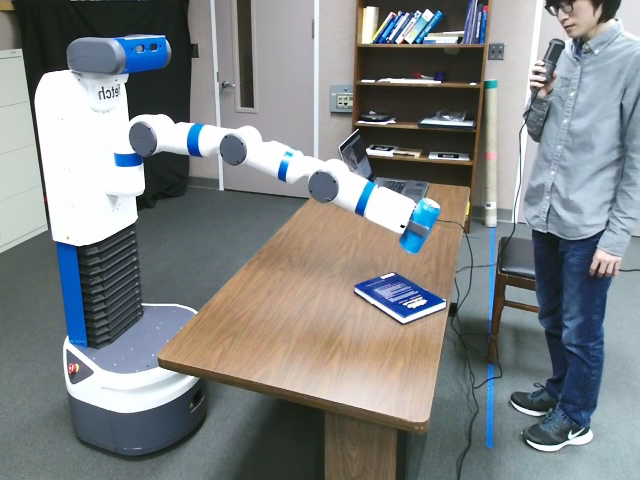}
  }
  \subfloat[][]
  {
    \includegraphics[width=0.32\linewidth]{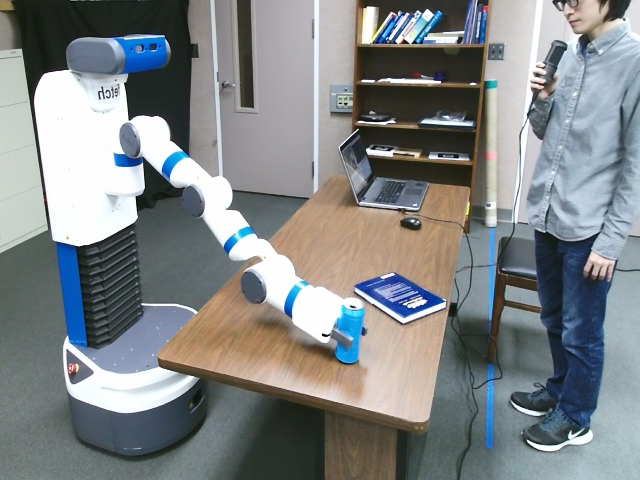}
  }
  \caption{The Fetch robot is moving a soda can on a table. Initially, the user gives the ``pick and place'' command. However, when the robot gets closer to the book, the person says \emph{``don't put it there''} and the robot avoids the book using appropriate cost functions and optimization.}
  \label{fig:stop}
\end{figure}

We also integrate our NLP-based planner with ROS and evaluated its performance on the 7-DOF Fetch robot. 
In a real-world setting, we test its performance  on different tasks corresponding to: (1)  moving a soda can on the table from one position to another; (2) not moving the soda can over the  book. With a noisy point cloud sensor on the robot, the thin book is not recognized as a separate obstacle by the robot, though the human user wants the robot to avoid it. All the instructions used in these tasks have different attributes, which makes it hard for prior methods. In Fig.~\ref{fig:dont}, the two sub-tasks are specified in one sentence at the beginning, as \emph{``move the can on the table, but don't put it on the book"}. The cost function is used to move the robot's end-effector to the surface of the table. Another cost function penalizes the distance between the book and the end-effector. In Fig.~\ref{fig:stop}, only the first sub-task is given at the beginning. This results in the robot moving the can on the book. As the robot gets too close to the book, the person says \emph{``stop,''} then says \emph{``don't put it there.''} The robot recomputes the cost functions and avoids the region around the book.

\subsection{Analysis}
We evaluate the performance based on the following metrics:

\noindent \emph{\bf Success Rate:} The ratio of successful task completion among all trials. Failure includes colliding with the obstacles due to an incorrect mapping of cost function parameters, violating constraints specified by natural language commands, and not completing the task due to some other reason.

\noindent \emph{\bf Trajectory Duration:} The time between the giving of the first NLP command and the robot's successful completion of the task after trajectory computation. A shorter duration implies a higher performance.

\noindent \emph{\bf Trajectory Smoothness Cost:} A cost based on evaluating the trajectory smoothness according to standard metrics and dividing it by the trajectory duration. A lower cost implies a smoother and more stable robot trajectory.

{\small
\begin{table}[t]
\caption{Planning performances with varying sizes of training data for the  scenario in Fig.~\ref{fig:sim_1} with 21 different NLP instructions.
}
\vspace*{-0.12in}
\centering
\scalebox{0.9}{
\begin{tabular}{|c|c|c|c|}
\hline
\# Training Data & Success Rate &  Duration &  Smoothness Cost \\ \hline
  1,000 &  5/10 & 23.46s (5.86s) & 8.72 (5.56) \\ \hline
  3,000 &  9/10 & 16.02s (3.28s) & 2.56 (0.64) \\ \hline
 10,000 & 10/10 & 13.16s (1.24s) & 1.21 (0.32) \\ \hline
 30,000 & 10/10 & 12.81s (0.99s) & 0.78 (0.12) \\ \hline
100,000 & 10/10 & 12.57s (0.97s) & 0.72 (0.10) \\ \hline
\end{tabular}
}
\label{table:training_data}
\end{table}
}
{\small
\begin{table}[t]
\caption{Running time (ms) of our DGG and motion planning modules for each scenario.
}
\vspace*{-0.12in}
\centering
\scalebox{0.9}{
\begin{tabular}{|c|c|c|c|c|}
\hline
Scenarios & Instructions & $|H|$ & \begin{tabular}[x]{@{}c@{}}DGG\\Time\end{tabular} & \begin{tabular}[x]{@{}c@{}}Planning\\Time\end{tabular} \\ \hline
Pick up an object (Fig. 4) & 10 & 12 & 32ms & 93ms \\ \hline
Don't put on the laptop (Fig. 5) & 20 & 13 & 16ms & 98ms \\ \hline
Move around obstacle & 45 & 9 & 16ms & 95ms \\ \hline
Static Instructions (video) & 20 & 18 & 73ms & 482ms \\ \hline
Dynamic Instructions (Fig. 1) & 21 & 18 & 58ms & 427ms \\ \hline
\end{tabular}
}
\label{table:time}
\vspace*{-0.2in}
\end{table}
}
Table~\ref{table:training_data} shows the results on our benchmarks with varying numbers of training data samples in the simulation environment shown in Fig.~\ref{fig:sim_1}. When the number of training data samples increases, the success rate also increases while the trajectory duration and the trajectory smoothness cost decrease. Table~\ref{table:time} shows the running time of our algorithm and the distances from the obstacle on the table in the real-world scenarios. We use 8 parallel threads for parallel trajectory search in the motion planning module.

Table~\ref{table:appendix_dgg} and~\ref{table:appendix_trajectory} show the examples of the dataset. In Table~\ref{table:appendix_dgg}, DGGs with word phrase nodes, grounding nodes and correspondence variables are shown. The correspondence variables of the graphs on the left column are all \emph{true}, and the groundings are matched correctly. Whereas, the correspondence variables on the right column are mixed with \emph{true} and \emph{false}. The groundings are not matched if the correspondence variable is \emph{false}. Many data samples are generated by flipping the correspondence variables between \emph{true} and \emph{false} to increase the accuracy of the DGG inference step. In Table~\ref{table:appendix_trajectory}, the latent variables are shown for the examples of the grounding graphs and the environment. The cost function weights and other necessary cost function parameters are manually set in the data generation program.

\section{ Benefits and Comparisons}
Most prior methods that combine NLP and motion planning have focused on understanding natural language instructions to compute robot motion for simple environments and constraints. Most of these methods are limited to navigation applications~\cite{oh2016integrated,chung2015performance,duvallet2016inferring} or simple settings~\cite{branavan2012learning}, or they are not evaluated on real robots~\cite{arkin2015towards}. Nyga et al.~\cite{nyga2012everything,nyga2018cloud,nyga2017instruction,nyga2017no} use probabilistic relation models based on knowledge bases to understand natural language commands that describe visual attributes of objects. This is complementary to our work. Broad et al.~\cite{broad2017real} extedd DCG for a robot manipulator so that it will handle natural language correction for robot motion in realtime. In our approach, the goal is to generate appropriate high-DOF motion trajectories in response to attribute-based natural language instructions like negation, distance or orientation constraints, etc. Unlike prior methods, the output of our NLP parsing algorithm is directly coupled with the specification of the motion planning problem as a constrained optimization method. 

It may be possible to extend prior methods~\cite{kollar2013generalized,howard2014natural} to handle attribute-based NLP instructions. For example, distance attributes require a number of constraints in the motion planning formulation. In natural language instructions such as \emph{``Pick up the blue block and put it 20 cm to the left of the red block''} or \emph{``Pick up one of the two blocks on the rightmost, and place it 10 inches away from the block on the leftmost,''} the exact distance specifications are the distance attributes. Prior methods that use G$^3$, DCG, or the Hybrid G$^3$-DCG models have only been evaluated with a small number of attributes (distance, orientation, and contact) to solve constrained motion planning problems. These prior techniques use discretized constraints~\cite{howard2014natural}, each of which can be active (i.e. $f(x) > 0)$), inverted ($f(x) < 0$), or ignored (i.e. not included). Therefore,  it is not possible to represent an explicit  constraint corresponding to the value of the  continuous variable \emph{distance}  in their formulation.

\section{Limitations, Conclusions and Future Work}
We present a motion planning algorithm that computes appropriate motion trajectories for a robot based on complex NLP instructions. Our formulation is based on two novel concepts: dynamic grounding graphs and dynamic constraint mapping. We highlight the performance in simulated and real-world scenes with a 7-DOF manipulator operating next to humans. We use a high dimensional optimization algorithm and the solver may get stuck in local minima, though we use multiple initializations to solve this problem. Furthermore, the accuracy of the mapping algorithm varies as a function of the training data.

As future work, we would like to overcome these limitations and evaluate the approach in challenging scenarios with moving obstacles while performing complex robot tasks. More work is needed to handle the full diversity of a natural language, especially for rare words, complicated grammar styles, and hidden intentions or emotions in human speech. We plan to incorporate stronger natural language processing and machine learning methods such as those based on semantic parsing, neural sequence-to-sequence models, etc. We also plan to collect more natural language data from a variety of sources such as recipes or demonstration videos.

\addtolength{\textheight}{-12cm}

\bibliographystyle{IEEEtran}
\bibliography{IEEEabrv,refs}

\clearpage
\begin{table*}[p]
\caption{Examples of DGGs with different configurations of correspondence variables.}
\centering
\begin{tabular}{|c|c|c|c|c|c|}
\hline
\multicolumn{6}{|c|}{Pick up one of the blue objects.} \\ \hline
Grounding Graph & \multicolumn{2}{|c|}{DGG Nodes} & Grounding Graph & \multicolumn{2}{|c|}{DGG Nodes} \\ \hline

% Case 1 & 2
\multirow{15}{*}{
  \includegraphics[width=0.23\textwidth]{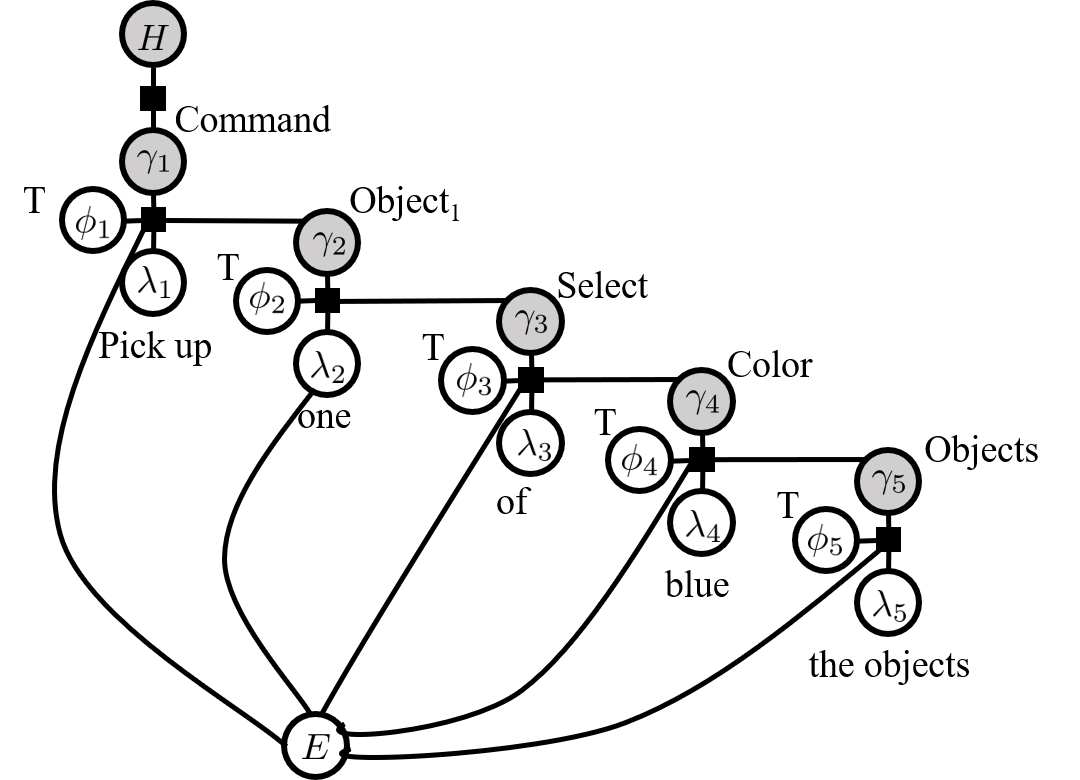}
}
& $\lambda_1$ & \emph{``Pick up''}
&
\multirow{15}{*}{
  \includegraphics[width=0.23\textwidth]{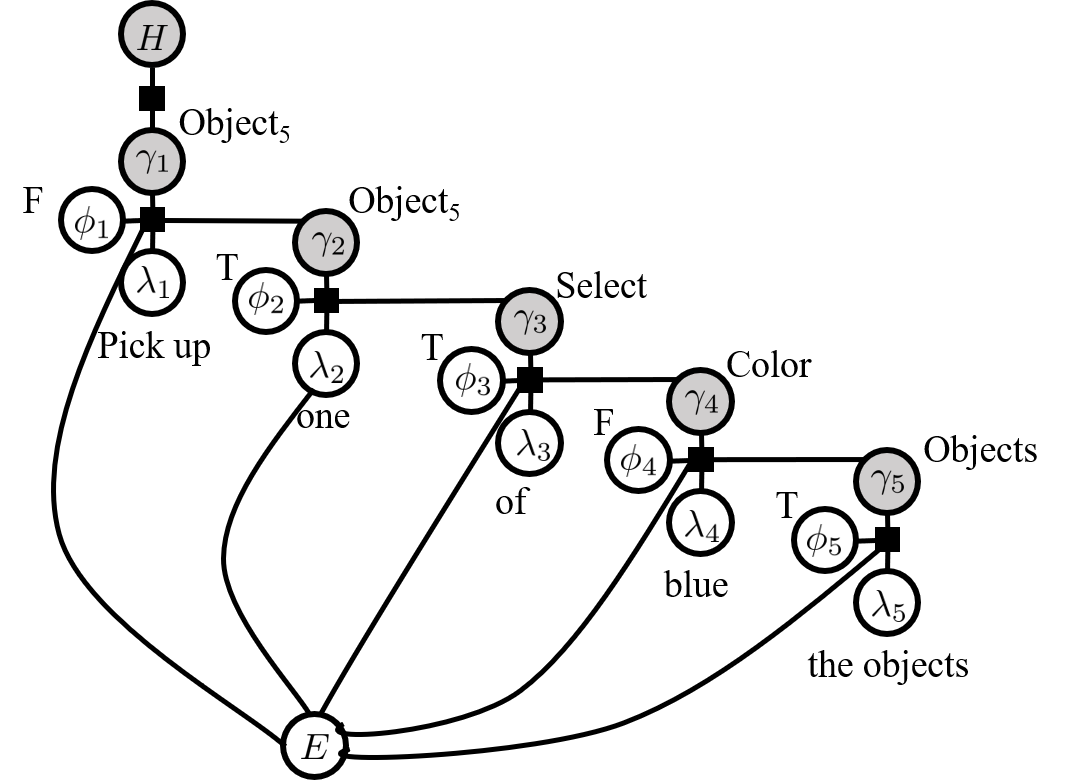}
}
& $\lambda_1$ & \emph{``Pick up''} \\ \cline{2-3} \cline{5-6}
& $\gamma_1$  & Command(pick up) & & $\gamma_1$ & Object$_5$ \\ \cline{2-3} \cline{5-6}
& $\phi_1$    & \emph{true}      & & $\phi_1$    & \emph{false} \\ \cline{2-3} \cline{5-6}
& $\lambda_2$ & \emph{``one''}   & & $\lambda_2$ & \emph{``one''} \\ \cline{2-3} \cline{5-6}
& $\gamma_2$  & Object$_1$       & & $\gamma_2$  & Object$_5$ \\ \cline{2-3} \cline{5-6}
& $\phi_2$    & \emph{true}      & & $\phi_2$    & \emph{true}\\ \cline{2-3} \cline{5-6}
& $\lambda_3$ & \emph{``of''}    & & $\lambda_3$ & \emph{``of''}\\ \cline{2-3} \cline{5-6}
& $\gamma_3$  & Select(nearest)  & & $\gamma_3$  & Select(nearest) \\ \cline{2-3} \cline{5-6}
& $\phi_3$    & \emph{true}      & & $\phi_3$    & \emph{true} \\ \cline{2-3} \cline{5-6}
& $\lambda_4$ & \emph{``blue''}  & & $\lambda_4$ & \emph{``blue''} \\ \cline{2-3} \cline{5-6}
& $\gamma_4$  & Color(blue)      & & $\gamma_4$  & Color(red) \\ \cline{2-3} \cline{5-6}
& $\phi_4$    & \emph{true}      & & $\phi_4$    & \emph{false}\\ \cline{2-3} \cline{5-6}
& $\lambda_5$ & \emph{``the objects''} & & $\lambda_5$ & \emph{``the objects''} \\ \cline{2-3} \cline{5-6}
& $\gamma_5$  & \{Object$_1$, $\cdots$, Object$_5$\} & & $\gamma_5$  & \{Object$_1$, $\cdots$, Object$_5$\} \\ \cline{2-3} \cline{5-6}
& $\phi_5$    & \emph{true} & & $\phi_5$    & \emph{true} \\ \hline

% Case 3 & 4
\multicolumn{6}{c}{} \\ \hline
\multicolumn{6}{|c|}{Place it on the table.} \\ \hline

\multirow{15}{*}{
  \includegraphics[width=0.2\textwidth]{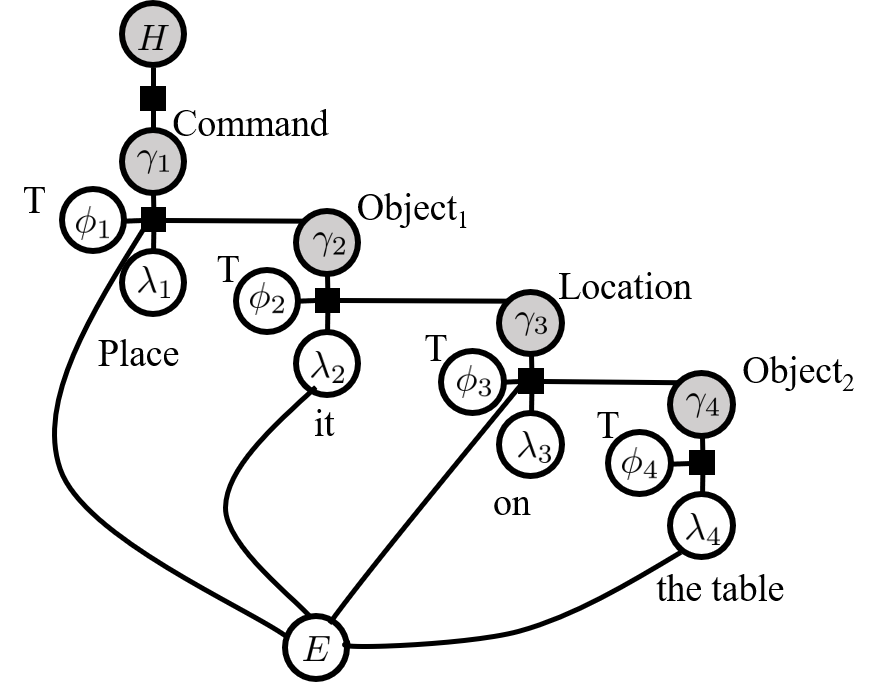}
}
& $\lambda_1$ & \emph{``Place''}
&
\multirow{15}{*}{
  \includegraphics[width=0.2\textwidth]{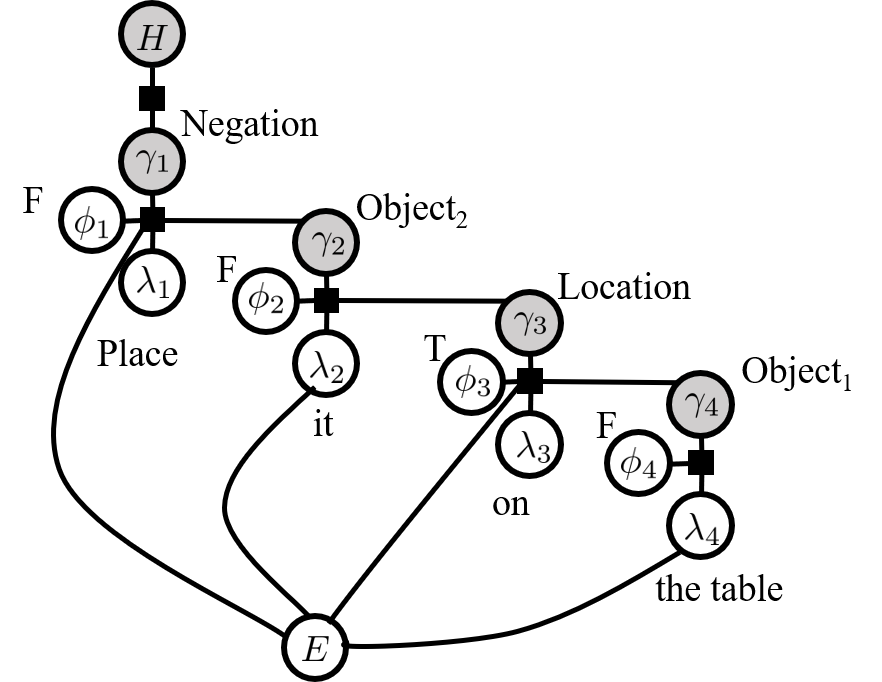}
}
& $\lambda_1$ & \emph{``Don't''} \\ \cline{2-3} \cline{5-6}
& $\gamma_1$  & Command(place) & & $\gamma_1$  & Negation \\ \cline{2-3} \cline{5-6}
& $\phi_1$    & \emph{true}    & & $\phi_1$    & \emph{false} \\ \cline{2-3} \cline{5-6}
& $\lambda_2$ & \emph{``it''}  & & $\lambda_2$ & \emph{``it''} \\ \cline{2-3} \cline{5-6}
& $\gamma_2$  & Object$_1$     & & $\gamma_2$  & Object$_2$ \\ \cline{2-3} \cline{5-6}
& $\phi_2$    & \emph{true}    & & $\phi_2$    & \emph{false} \\ \cline{2-3} \cline{5-6}
& $\lambda_3$ & \emph{``on''}  & & $\lambda_3$ & \emph{``on''} \\ \cline{2-3} \cline{5-6}
& $\gamma_3$  & Location(on)   & & $\gamma_3$  & Location(on) \\ \cline{2-3} \cline{5-6}
& $\phi_3$    & \emph{true}    & & $\phi_3$    & \emph{true} \\ \cline{2-3} \cline{5-6}
& $\lambda_4$ & \emph{``the table''} & & $\lambda_4$ & \emph{``the table''}\\ \cline{2-3} \cline{5-6}
& $\gamma_4$  & Object$_2$           & & $\gamma_4$  & Object$_1$ \\ \cline{2-3} \cline{5-6}
& $\phi_4$    & \emph{true}          & & $\phi_4$    & \emph{false} \\ \cline{2-3} \cline{5-6}
& \multicolumn{2}{c|}{} & & \multicolumn{2}{c|}{} \\
& \multicolumn{2}{c|}{} & & \multicolumn{2}{c|}{} \\
& \multicolumn{2}{c|}{} & & \multicolumn{2}{c|}{} \\ \hline

% Case 5 & 6
\multicolumn{6}{c}{} \\ \hline
\multicolumn{6}{|c|}{Don't put it there.} \\ \hline

\multirow{15}{*}{
  \includegraphics[width=0.23\textwidth]{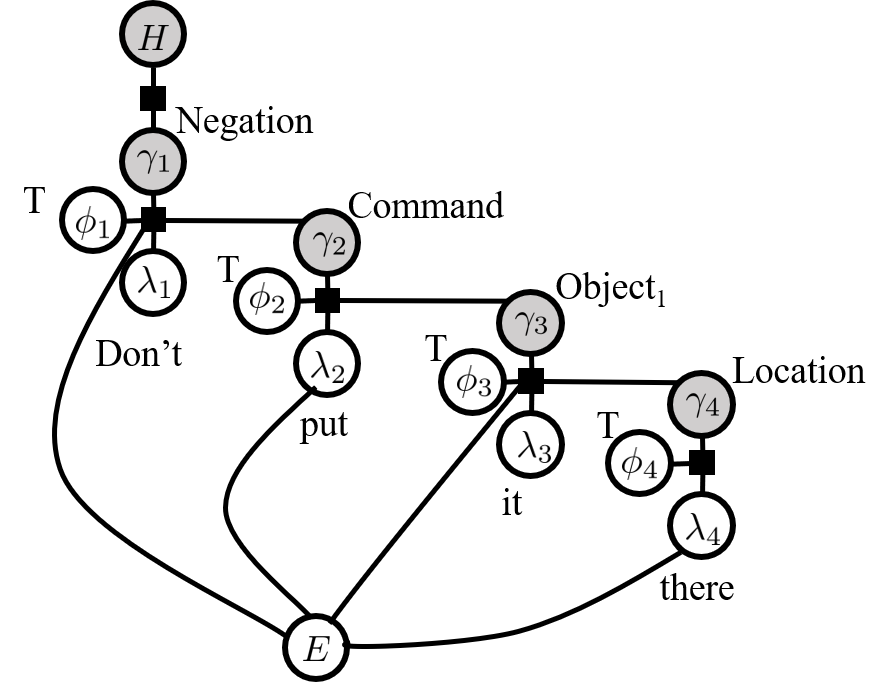}
}
& $\lambda_1$ & \emph{``Place''}
&
\multirow{15}{*}{
  \includegraphics[width=0.23\textwidth]{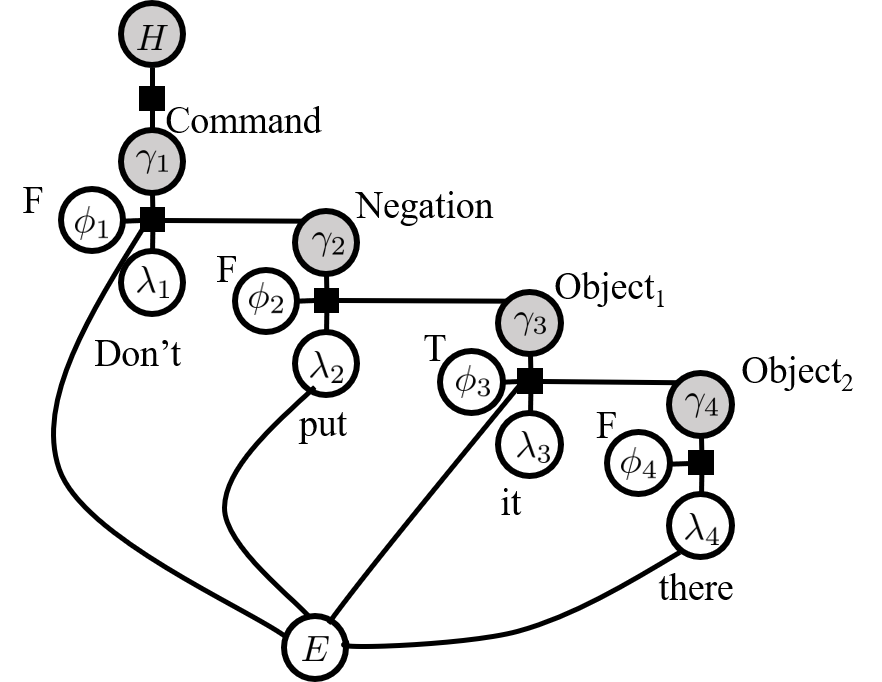}
}
& $\lambda_1$ & \emph{``Place''} \\ \cline{2-3} \cline{5-6}
& $\gamma_1$  & Negation         & & $\gamma_1$  & Command(place) \\ \cline{2-3} \cline{5-6}
& $\phi_1$    & \emph{true}      & & $\phi_1$    & \emph{false} \\ \cline{2-3} \cline{5-6}
& $\lambda_2$ & \emph{``put''}   & & $\lambda_2$ & \emph{``put''} \\ \cline{2-3} \cline{5-6}
& $\gamma_2$  & Command(place)   & & $\gamma_2$  & Negation \\ \cline{2-3} \cline{5-6}
& $\phi_2$    & \emph{true}      & & $\phi_2$    & \emph{false} \\ \cline{2-3} \cline{5-6}
& $\lambda_3$ & \emph{``it''}    & & $\lambda_3$ & \emph{``it''} \\ \cline{2-3} \cline{5-6}
& $\gamma_3$  & Object$_1$       & & $\gamma_3$  & Object$_1$ \\ \cline{2-3} \cline{5-6}
& $\phi_3$    & \emph{true}      & & $\phi_3$    & \emph{true} \\ \cline{2-3} \cline{5-6}
& $\lambda_4$ & \emph{``there''} & & $\lambda_4$ & \emph{``there''}\\ \cline{2-3} \cline{5-6}
& $\gamma_4$  & Location(robot)  & & $\gamma_4$  & Object$_2$ \\ \cline{2-3} \cline{5-6}
& $\phi_4$    & \emph{true}      & & $\phi_4$    & \emph{false} \\ \cline{2-3} \cline{5-6}
& \multicolumn{2}{c|}{} & & \multicolumn{2}{c|}{} \\
& \multicolumn{2}{c|}{} & & \multicolumn{2}{c|}{} \\
& \multicolumn{2}{c|}{} & & \multicolumn{2}{c|}{} \\ \hline
\end{tabular}
\label{table:appendix_dgg}
\end{table*}

\begin{table*}[p]
\caption{Examples of DGGs and the latent variables.}
\centering
\begin{tabular}{|c|c|c|c|}
\hline
\multicolumn{4}{|c|}{Pick up one of the blue objects} \\ \hline

% Case 1
Grounding Graph & \multicolumn{2}{|c|}{DGG Nodes} & Trajectory \\ \hline
\multirow{14}{*}{
  \includegraphics[width=0.27\textwidth]{figs/app_dgg1.png}
}
& $\gamma_1$ & Command(pick up) & 
\multirow{14}{*}{
  \includegraphics[width=0.27\textwidth]{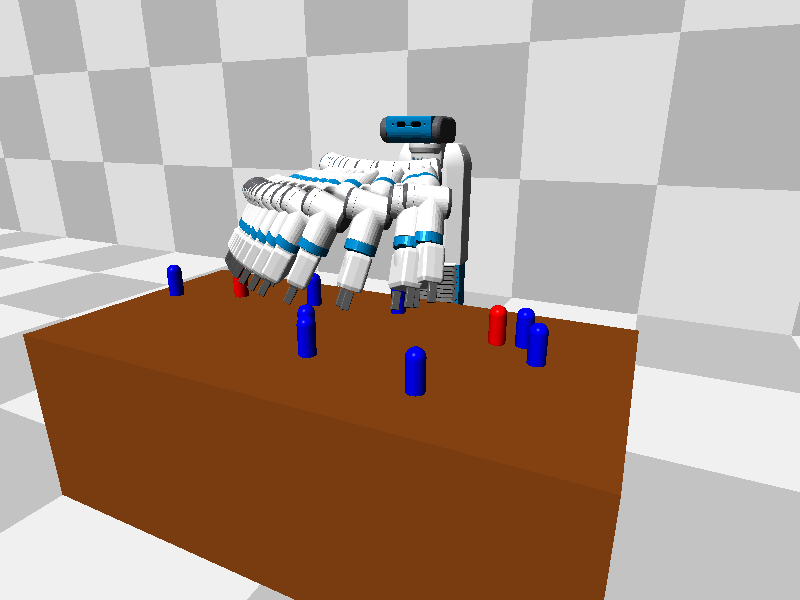}
}
\\ \cline{2-3}
& $\gamma_2$ & Object$_1$ & \\ \cline{2-3}
& $\gamma_3$ & Select(nearest) & \\ \cline{2-3}
& $\gamma_4$ & Color(blue) & \\ \cline{2-3}
& $\gamma_5$ & \{Object$_1$, $\cdots$, Object$_5$\} & \\ \cline{2-3}
& \multicolumn{2}{|c|}{Latent Variables} & \\ \cline{2-3}
& Collision avoidance & 3.00 & \\ \cline{2-3}
& Smoothness          & 1.00 & \\ \cline{2-3}
& \multirow{2}{*}{End-effector position} & 10.0 & \\ \cline{3-3}
& & (0.75, 0.2, 0.81) & \\ \cline{2-3}
& \multirow{2}{*}{End-effector orientation} & 30.00 & \\ \cline{3-3}
& & (0.00, 0.00, 1.00) & \\ \cline{2-3}
& End-effector speed       & 0.00 & \\ \cline{2-3}
& Repulsion                & 0.00 & \\ \hline

% Case 2
Grounding Graph & \multicolumn{2}{|c|}{DGG Nodes} & Trajectory \\ \hline
\multirow{14}{*}{
  \includegraphics[width=0.27\textwidth]{figs/app_dgg1.png}
}
& $\gamma_1$ & Command(pick up) & 
\multirow{14}{*}{
  \includegraphics[width=0.27\textwidth]{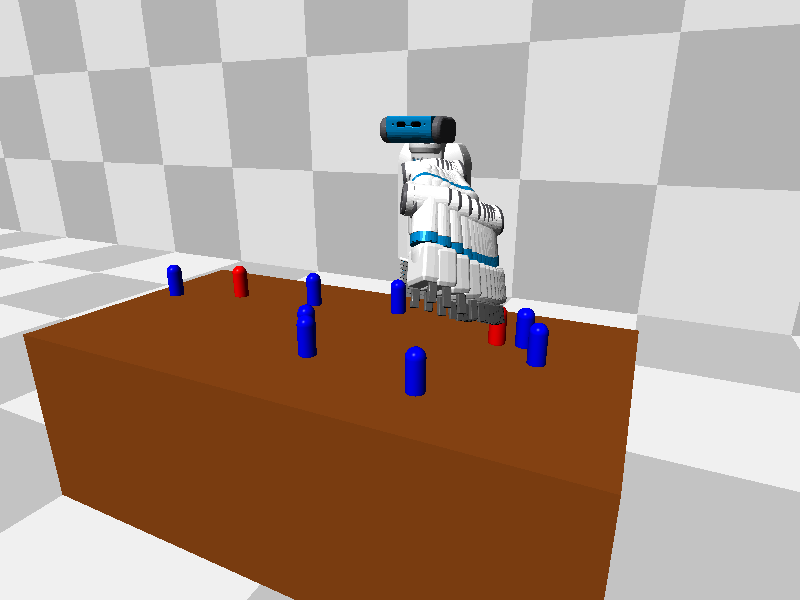}
}
\\ \cline{2-3}
& $\gamma_2$ & Object$_1$ & \\ \cline{2-3}
& $\gamma_3$ & Select(nearest) & \\ \cline{2-3}
& $\gamma_4$ & Color(blue) & \\ \cline{2-3}
& $\gamma_5$ & \{Object$_1$, $\cdots$, Object$_5$\} & \\ \cline{2-3}
& \multicolumn{2}{|c|}{Latent Variables} & \\ \cline{2-3}
& Collision avoidance & 1.00 & \\ \cline{2-3}
& Smoothness          & 3.00 & \\ \cline{2-3}
& \multirow{2}{*}{End-effector position} & 10.0 & \\ \cline{3-3}
& & (-0.43, 0.26, 0.81) & \\ \cline{2-3}
& \multirow{2}{*}{End-effector orientation} & 30.00 & \\ \cline{3-3}
& & (0.00, 0.00, 1.00) & \\ \cline{2-3}
& End-effector speed       & 0.00 & \\ \cline{2-3}
& Repulsion                & 0.00 & \\ \hline

\multicolumn{4}{c}{} \\ \hline

\multicolumn{4}{|c|}{Place it on the table.} \\ \hline

% Case 3
Grounding Graph & \multicolumn{2}{|c|}{DGG Nodes} & Trajectory \\ \hline
\multirow{14}{*}{
  \includegraphics[width=0.24\textwidth]{figs/app_dgg3.png}
}
& $\gamma_1$ & Command(place) & 
\multirow{14}{*}{
  \includegraphics[width=0.27\textwidth]{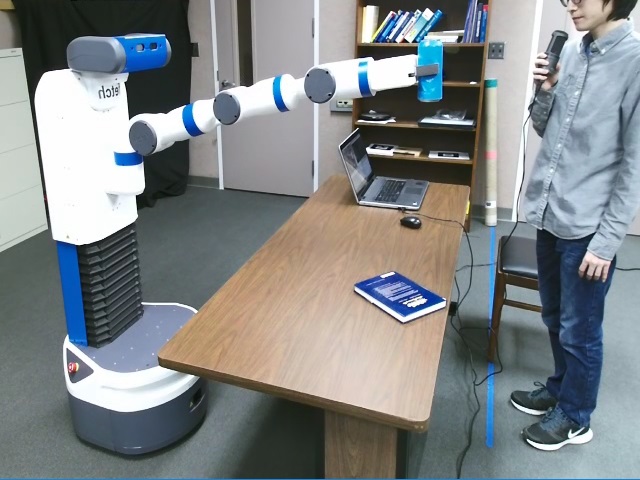}
}
\\ \cline{2-3}
& $\gamma_2$ & Object$_1$ & \\ \cline{2-3}
& $\gamma_3$ & Location(on) & \\ \cline{2-3}
& $\gamma_4$ & Object$_2$ & \\ \cline{2-3}
& \multicolumn{2}{|c|}{} & \\ \cline{2-3}
& \multicolumn{2}{|c|}{Latent Variables} & \\ \cline{2-3}
& Collision avoidance & 1.00 & \\ \cline{2-3}
& Smoothness          & 3.00 & \\ \cline{2-3}
& \multirow{2}{*}{End-effector position} & 10.0 & \\ \cline{3-3}
& & (0.00, 0.30, 0.70) & \\ \cline{2-3}
& \multirow{2}{*}{End-effector orientation} & 100 & \\ \cline{3-3}
& & (0.00, 0.00, 1.00) & \\ \cline{2-3}
& End-effector speed & 0.00 & \\ \cline{2-3}
& Repulsion & 0.00 & \\ \hline

\multicolumn{4}{c}{} \\ \hline

\multicolumn{4}{|c|}{Don't put it there.} \\ \hline

% Case 4
Grounding Graph & \multicolumn{2}{|c|}{DGG Nodes} & Trajectory \\ \hline
\multirow{15}{*}{
  \includegraphics[width=0.24\textwidth]{figs/app_dgg5.png}
}
& $\gamma_1$ & Negation & 
\multirow{15}{*}{
  \includegraphics[width=0.27\textwidth]{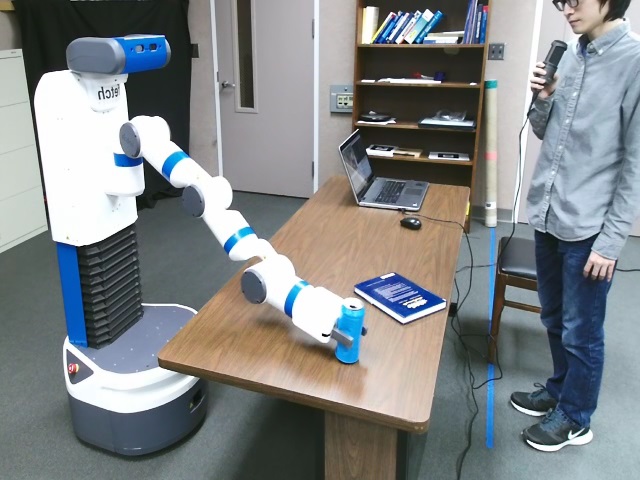}
}
\\ \cline{2-3}
& $\gamma_2$ & Command(place) & \\ \cline{2-3}
& $\gamma_3$ & Object$_1$ & \\ \cline{2-3}
& $\gamma_4$ & Location(robot) & \\ \cline{2-3}
& \multicolumn{2}{|c|}{} & \\ \cline{2-3}
& \multicolumn{2}{|c|}{Latent Variables} & \\ \cline{2-3}
& Collision avoidance & 1.00 & \\ \cline{2-3}
& Smoothness & 3.00 & \\ \cline{2-3}
& \multirow{2}{*}{End-effector position} & 10.0 & \\ \cline{3-3}
& & (0.00, 0.30, 0.70) & \\ \cline{2-3}
& \multirow{2}{*}{End-effector orientation} & 100 & \\ \cline{3-3}
& & (0.00, 0.00, 1.00) & \\ \cline{2-3}
& End-effector speed & 0.00 & \\ \cline{2-3}
& \multirow{2}{*}{Repulsion} & 3.00 & \\ \cline{3-3}
& & 10.00, (0.05, 0.32, 0.70) & \\ \hline

\end{tabular}
\label{table:appendix_trajectory}
\end{table*}

\end{document}